\documentclass[letterpaper]{article} 
\usepackage{sty/aaai2027}
\usepackage[hyphens]{url}  
\usepackage{graphicx} 
\usepackage{natbib}  
\usepackage{caption} 
\usepackage[T1]{fontenc}
\usepackage{ifthen}
\usepackage{algorithm}
\usepackage[noEnd=true,indLines=true,commentColor=black]{sty/algpseudocodex}
\usepackage{amsmath,amssymb,amsthm}
\usepackage{booktabs}
\usepackage{subcaption}
\usepackage{cleveref}
\usepackage{siunitx}
\usepackage{makecell}
\usepackage{multirow}
\usepackage{sty/custom}

\nocopyright

\usepackage{newfloat}
\usepackage{listings}
\DeclareCaptionStyle{ruled}{labelfont=normalfont,labelsep=colon,strut=off} 
\floatstyle{ruled}
\newfloat{listing}{tb}{lst}{}
\floatname{listing}{Listing}
\title{Travel Cost Calibration for Multi-Agent Target Assignment and Pathfinding}
\author{Yu Kumagai$^{1,2}$, Ryohei Banno$^{1}$, Keisuke Okumura$^{2}$}
\affiliations{
  $^1$Hitotsubashi University, Japan
  \\
  $^2$National Institute of Advanced Industrial Science and Technology (AIST), Japan
  \\
  {yu.kumagai@aist.go.jp, r.banno@r.hit-u.ac.jp, okumura.k@aist.go.jp}
}

\begin{document}

\maketitle

\begin{abstract}
Target assignment and pathfinding (TAPF) is an appealing extension of classical multi-agent pathfinding (MAPF): each agent has a set of targets, and the planner must both assign a target to each agent and plan collision-free paths for them.
A trivial approach is to compute an assignment that minimizes the total single-agent shortest-path distance over agent--target pairs, and then solve the resulting MAPF instance.
However, this approach falls short because the shortest-path distance remains a proxy that ignores interference among agents.
Assigning targets based on the costs agents actually pay, which depend on how every other agent is assigned and routed, would yield a better final outcome.
To this end, we propose an iterative TAPF method that improves target assignments by calibrating travel cost estimates from fast, suboptimal MAPF feedback.
Experiments show that our method produces solutions close to optimal on small instances and outperforms existing planners in solution quality on larger ones.
\end{abstract}
\section{Introduction}
Multi-agent pathfinding (MAPF)~\cite{stern2019def}, the problem of finding
collision-free paths that bring each agent to its own pre-assigned target on a shared graph, is a core component of multi-agent systems such as automated warehouses.
While finding an optimal solution is known to be computationally intractable~\cite{yu2013structure}, the past decade has produced mature fast suboptimal solvers---notably PIBT~\cite{okumura2022priority} and LaCAM~\cite{okumura2023lacam}---that reliably plan the paths for hundreds to thousands of agents in real time.

In real-world logistics deployments, however, tasks are often \emph{partially agent-agnostic}.
Each agent has a set of possible targets, and the planner must both decide which target each agent goes to and plan a collision-free path.
This combined problem is known as \emph{Target Assignment and PathFinding (TAPF)}~\cite{ma2016tapf}.

A trivial approach to TAPF is two-step planning;
it first builds an agent-target cost matrix from single-agent shortest-path distances, applies a linear optimal assignment to minimize the total cost, and then plans collision-free paths for the resulting MAPF instance.
However, this approach falls short because the shortest-path distances ignore the interference between agents.
\begin{figure}[t!]
    \centering
    \begin{minipage}[c]{0.035\columnwidth}
        \centering
        \rotatebox[origin=c]{90}{\footnotesize $\leftarrow$ flowtime / LB}
    \end{minipage}\hfill
    \begin{minipage}[c]{0.235\columnwidth}
        \centering
        \includegraphics[width=\linewidth]{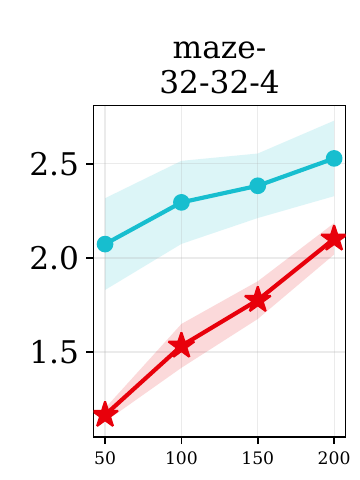}
    \end{minipage}\hfill
    \begin{minipage}[c]{0.235\columnwidth}
        \centering
        \includegraphics[width=\linewidth]{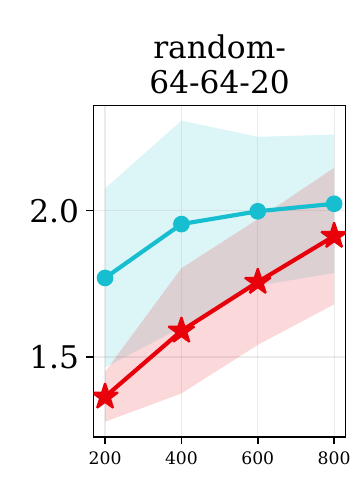}
    \end{minipage}\hfill
    \begin{minipage}[c]{0.235\columnwidth}
        \centering
        \includegraphics[width=\linewidth]{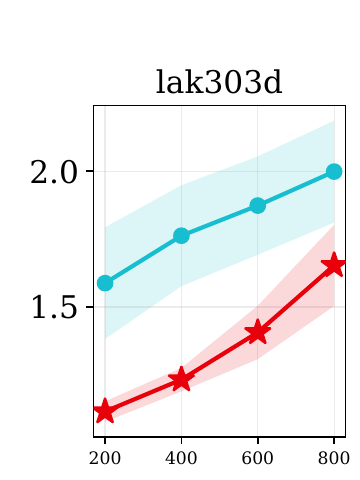}
    \end{minipage}\hfill
    \begin{minipage}[c]{0.235\columnwidth}
        \centering
        \includegraphics[width=\linewidth]{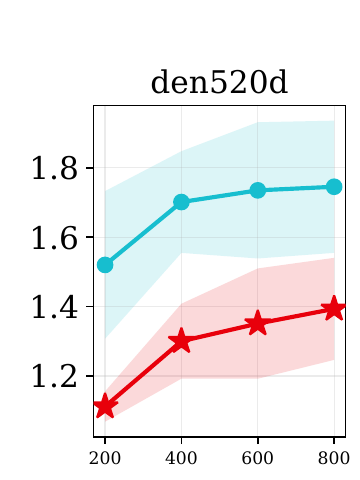}
    \end{minipage}\\[1pt]
    \hspace*{0.035\columnwidth}{\footnotesize agents}\\[3pt]
    \includegraphics[width=\columnwidth]{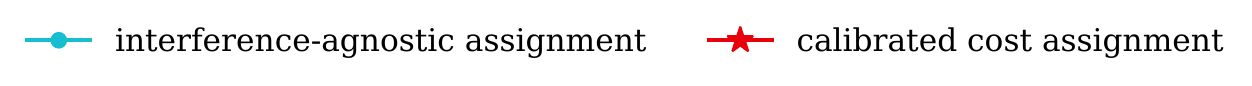}
    \caption{
    TAPF solution quality of an interference-agnostic assignment and a calibrated cost assignment in the \scenHotspot scenario, where the targets are densely clustered in small regions of the map.
    Targets are assigned by a linear sum assignment on each, and paths are planned by LaCAM3~\cite{okumura2024lacamstar}, a fast suboptimal MAPF solver.
    }
    \label{fig:initial_vs_calibrated}
\end{figure}
If the travel costs that the agents actually pay while interacting with one another were available, the best assignment would follow from a single linear sum assignment on those costs.
Observing such costs, however, poses a chicken-and-egg problem.
The actual travel cost is revealed only by the planned paths, and the planned paths change with the target assignment.
A cost matrix filled with the actual travel costs of every pair therefore does not exist as a fixed object.
Nevertheless, as \cref{fig:initial_vs_calibrated} shows, a target assignment on costs calibrated toward the actual travel costs yields a far better solution than an interference-agnostic assignment.

This paper provides a concrete approach for this calibration idea.
We propose \emph{Target Assignment with Calibrated Travel costs (\planner)}, an algorithm that solves TAPF by gradually calibrating the travel cost estimates of agent-target pairs using fast MAPF feedback.
Specifically, \planner plans paths for diverse target assignments with a fast suboptimal MAPF solver and feeds the observed travel costs back, so that the estimates progressively reflect what the agents actually pay. 

We compare \planner with representative TAPF solvers on standard MAPF benchmark maps~\cite{stern2019def}.
While \planner offers no guarantee on solution quality, it stays near-optimal on small instances, on par with ITA-ECBS~\cite{tang2024ita-ecbs}, a state-of-the-art bounded-suboptimal TAPF solver.
On large instances with up to 800 agents, which lie beyond the reach of ITA-ECBS within realistic time budgets, \planner attains better solution quality than DBS-Hungarian~\cite{kumagai2026alternating}, an iterative refinement solver designed for large-scale TAPF.
Furthermore, \planner is scalable; it can handle massive-scale TAPF instances with $10{,}000$ agents.
Together, these results push the frontier of solution quality in large-scale TAPF.
The code is available at \url{https://github.com/Ukuma012/TACT}.

\section{Preliminaries}
\subsection{Problem Definition}
A TAPF instance is defined on an undirected graph $G=(V, E)$ shared by a set of agents $A=\{1, \ldots, n\}$.
Each agent $i \in A$ starts from a vertex $s_i \in V$, and its admissible targets are specified by a binary matrix $\mathcal{T}$ of dimension $n \times |V|$: the entry $\mathcal{T}[i][v] = 1$ marks vertex $v \in V$ as an eligible target of agent $i$.
We require $\mathcal{T}$ to admit at least one feasible assignment, where an \emph{assignment} is an injective mapping $\mathcal{M}: A \to V$ for which $\mathcal{T}[i][\mathcal{M}(i)] = 1$ for every agent $i$.

Solving the instance means producing such an assignment $\assignment$ together with a set of collision-free paths $\Pi = \{\pi_1, \ldots, \pi_n\}$, where each path $\pi_i$ starts at $s_i$ and ends at $\assignment(i)$.
We follow the standard MAPF model of motion and collisions~\cite{stern2019def}.
Time proceeds in discrete steps, and at each step an agent either moves to an adjacent vertex or waits at its current one.
Two types of conflict are prohibited: a \emph{vertex conflict}, where two agents occupy the same vertex at the same step, and an \emph{edge conflict}, where two agents traverse the same edge in opposite directions.

We use \emph{travel cost} as a generic term for the cost an agent pays to reach its target.
Three concrete notions of it appear throughout the paper.
\begin{itemize}
    \item The \emph{shortest-path distance} $\dist(s_i, g)$ is the length of a shortest path from the start $s_i$ of agent $i$ to a target $g$, ignoring all the other agents.
    \item The \emph{travel loss} $c_i(\Pi)$ of agent $i$ is the number of time steps at which it is not resting at its target $\assignment(i)$.
    \item The \emph{travel time} $t_i(\Pi)$ of agent $i$ is the number of time steps it takes until it reaches its target and stops moving.
    The \emph{flowtime} is the sum of travel times over all agents, also known as sum-of-costs.
\end{itemize}
Our objective is to minimize the flowtime within a planning-time budget, following common practice in MAPF studies.
Note that \planner calibrates its cost matrix estimates with the \emph{travel loss} rather than the travel time.\footnote{This is because calibrating with the travel time leads to worse final solutions on some maps. See the appendix for the comparison.}

\subsection{Cost Matrix}
\label{sec:cost-matrix}
Assigning targets requires an estimate of the travel cost for every agent--target pair.
We collect these travel costs into a \emph{cost matrix} $C$ of dimension $n \times |V|$, where $C[i][v]$ is the cost of assigning vertex $v$ to agent $i$.
A natural choice for what $C$ encodes is $C[i][v] = \dist(s_i, v)$, but these distances ignore the interference between agents.
The ideal entry would instead be the travel loss $c_i(\Pi)$ that agent $i$ actually pays when assigned to $v$.
This quantity, however, is defined only with respect to a full solution, because the interference that agent $i$ experiences is shaped by the assignments and paths of all the other agents.
Different entries of $C$ presuppose different assignments, and no single solution realizes them all at once.
There is thus no fixed matrix of actual travel costs to be computed.
\planner therefore treats $C$ as an estimate, initialized with the shortest-path distances and calibrated with the travel losses observed in planned solutions.

\subsection{Linear Sum Assignment}
\label{sec:prelim-lap}
The \emph{linear sum assignment} asks for a one-to-one matching between two sets that minimizes the total cost of the matched pairs.
In our setting, it matches the agents to the vertices, with the pair costs given by a cost matrix $C$.
The problem is solvable in polynomial time, for instance by the Hungarian algorithm~\cite{kuhn1955hungarian} or, as we adopt, by the successive-shortest-path algorithm~\cite{ahuja1988network}.
Throughout this paper, the matching is additionally required to respect the target matrix, that is, every agent is matched to one of its candidate targets.
We call such a matching a \emph{target assignment} in the remainder of the paper.

\subsection{Related Work}
\paragraph{Utilization of Fast Suboptimal MAPF.}
Early research on MAPF centered on optimal solvers, most notably CBS~\cite{sharon2015cbs}, which guarantees optimal solutions but whose scalability is inherently limited.
A major recent advance is the emergence of fast solvers such as PIBT~\cite{okumura2022priority} and LaCAM~\cite{okumura2023lacam}, which reliably deliver solutions for thousands of agents in real time, although their solutions are suboptimal.
Anytime variants such as LaCAM3~\cite{okumura2024lacamstar} keep refining the solution as long as time permits, trading planning time for solution quality.

Fast suboptimal planners have opened up a new planning paradigm, where planning is no longer a one-shot computation.
Their short planning time allows a solver to be invoked repeatedly.
For example, PIE~\cite{zhang2024pie} aims to minimize the total of planning and execution times, so it starts execution as soon as a fast solver commits the first few actions and keeps improving the remaining paths while the agents move.
LF~\cite{shankar2025lf} targets online navigation of real robots, embedding repeated replanning by a fast solver in a control loop so that the robots adapt to goal updates and dynamic workspaces.
Another use of the repeated invocation, which we follow, is to improve the solution quality with feedback obtained from each iteration.
Lightweight Traffic Map~\cite{shen2026ltm} targets dense settings where conventional anytime refinement struggles.
By running LaCAM repeatedly and biasing each new plan with the congestion observed in the previous solution, it becomes a strong anytime MAPF solver.
DBS-Hungarian~\cite{kumagai2026alternating} brings this pattern to TAPF, which we describe in detail later.
Our study builds on the same principle and directs the feedback toward target assignment.

\paragraph{TAPF.}
Historically, most TAPF solvers have been developed as extensions of CBS.
These solvers all maintain a cost matrix, and they differ in what its entries hold.
CBS-TA~\cite{honig2018ecbs-ta} fills the cost matrix with shortest-path distances and searches a forest of constraint trees, one per assignment.
ITA-CBS~\cite{tang2023ita-cbs} and its bounded-suboptimal variant ITA-ECBS~\cite{tang2024ita-ecbs} keep a single constraint tree and refresh the cost matrix during the search, but only with each agent's constrained shortest-path distance.
In contrast, our approach updates the cost matrix entries with the travel costs that a MAPF solver observes, so the assignment accounts for the interference between agents that the shortest-path distances ignore.

While CBS-based approaches provide solution quality guarantees, they struggle to handle large-scale instances because the number of conflicts grows and the constraint tree expands to resolve them. 
DBS-Hungarian~\cite{kumagai2026alternating} instead interleaves target assignment with path planning by a fast suboptimal MAPF solver, repeatedly revising the assignment based on the paths just planned.
Specifically, in each iteration, it selects a subgroup of agents whose path costs are delayed behind their shortest-path distance, and re-solves a target assignment for the subgroup with the Hungarian method.
This design attains scalability to thousands of agents, but at the price of solution quality.
Our approach also exploits feedback from a fast suboptimal MAPF solver, but uses the feedback differently.
DBS-Hungarian performs a local search that reassigns targets within a small subgroup of agents, whereas \planner accumulates every observation into the cost matrix and revises the assignment of all agents at once.  

\paragraph{Optimization with Proxy Evaluations.}
Our approach to TAPF has a close connection to surrogate-based optimization~\cite{forrester2009surrogate}.
It maintains a computationally cheap proxy for the objective, optimizes this proxy, and refines it with the outcome of each evaluation.
Multifidelity methods~\cite{peherstorfer2018survey} go one step further and replace the expensive evaluation itself with a fast low-fidelity solver, accepting some error in exchange for many more evaluations.
Our method combines both ideas.
The cost matrix is a surrogate whose target assignment proposes the next candidate, and the fast suboptimal MAPF solver acts as the low-fidelity evaluator whose observed travel costs calibrate the cost matrix.

\section{\planner}
\label{sec:algorithm}
\begin{figure*}[t!]
        \includegraphics[width=\linewidth]{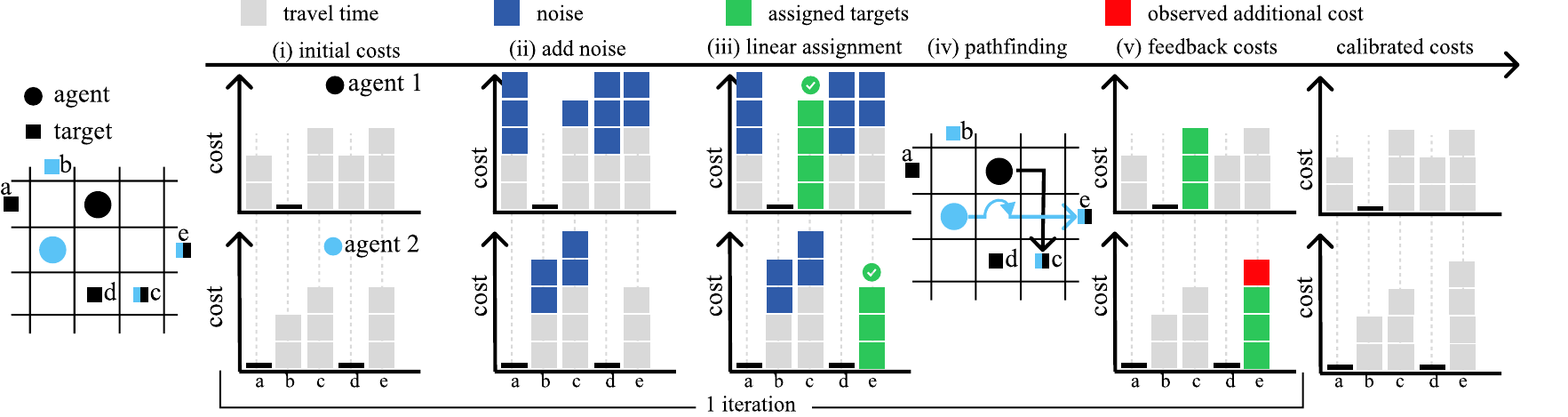}
    \caption{
    Overview of \planner.
    \emph{(i)}~The cost matrix is initialized with shortest-path distances, \emph{(ii, iii)}~Random noise varies the target assignment.
    Then, \emph{(iv)}~MAPF plans paths for it, and \emph{(v)}~the observed travel losses are fed back, gradually calibrating the estimates.
    \emph{(ii--iv)}~can be executed in parallel, so each iteration feeds back several observations.
    }
    \label{fig:overview}
\end{figure*}
{
\begin{algorithm}[t!]
\caption{TACT}
\label{algo:framework}
\begin{algorithmic}[1]
\small
\Input{Instance $I$, noise $\sigma$, \#samples $P$, learning rate $\alpha$}
\Output{task assignment $\assignment$ and collision-free paths $\solutions$}
\State $\CostMatrix \leftarrow \funcname{InitCost}(I)$,\;
$c\sub{best} \leftarrow \infty$,\;
$\assignment\sub{best} \leftarrow \bot$,\;
$\solutions\sub{best} \leftarrow \bot$
\label{algo:framework:init}

\While{not interrupted}
\For{$k = 1, \ldots, P$ \textbf{in parallel}}
\label{algo:framework:parallel}
\State $\CostMatrix'_k \leftarrow \CostMatrix + \sigma E_k$,
where $E_k$ has i.i.d.\ $\mathcal{U}(0, 1)$ entries
\label{algo:framework:noise}
\State $\assignment_k \leftarrow \funcname{LinearSumAssignment}(\CostMatrix'_k)$
\label{algo:framework:linear_assignment_with_noise}
\State $\solutions_k \leftarrow \funcname{SolveMAPF}(\assignment_k)$
\label{algo:framework:mapf}
\State $c_k \leftarrow \funcname{GetCost}(\solutions_k)$
\label{algo:framework:getcost}
\EndFor
\State $k^\ast \leftarrow \arg\min_k c_k$
\label{algo:framework:argmin}
\If{$c_{k^\ast} < c\sub{best}$}
\State $(\assignment\sub{best},\, c\sub{best},\, \solutions\sub{best})
 \leftarrow (\assignment_{k^\ast},\, c_{k^\ast},\, \solutions_{k^\ast})$
\label{algo:framework:track}
\EndIf
\State $\CostMatrix \leftarrow \funcname{UpdateMatrix}(\CostMatrix, \{(\assignment_k, \solutions_k)\}_{k=1}^{P}, \alpha)$
\label{algo:framework:update}
\EndWhile
\State $\solutions\sub{best} \leftarrow \funcname{SolveNearOptimalMAPF}(\assignment\sub{best})$
\Comment{optional}
\label{algo:framework:path-opt}
\State \Return $\assignment\sub{best}$, $\solutions\sub{best}$
\end{algorithmic}
\end{algorithm}
}
The proposed \planner solves TAPF by calibrating the travel cost of agent--target pairs through repeated path planning feedback with a fast suboptimal MAPF solver.
As illustrated in \cref{fig:overview}, \planner first initializes the travel cost estimate of each pair with the shortest-path distance.
Then, it repeats an iteration that consists of three steps.
\begin{enumerate}
    \item It adds noise to the travel cost estimates.
    \item It computes a target assignment on the perturbed estimates and plans paths for it with the MAPF solver.
    \item It feeds the travel losses observed in the planned paths back into the estimates.
\end{enumerate}
Through this iteration, the estimates are gradually calibrated toward the travel costs that agents actually pay.
\Cref{algo:framework} formalizes this idea.
In what follows, we describe how the noise produces diverse candidates and how the feedback updates the estimates.
\planner is simple to implement and also offers a practical advantage owing to its modularity.
Because its components are decoupled, future advances in the underlying MAPF planners (e.g., guidance~\cite{zhang2024guidance,arita2026local} or neural heuristics~\cite{jain2026graph}) can be integrated without modifications.

\subsection{Algorithm}
\paragraph{Initialization (\cref{algo:framework:init}).}
\planner stores its travel cost estimates in a matrix $\CostMatrix$.
For every feasible pair with $\mathcal{T}[i][v]=1$, \funcname{InitCost} sets the single-agent shortest-path distance.

\paragraph{Target Assignment Generation (\crefrange{algo:framework:parallel}{algo:framework:getcost}).}
The calibration needs observations, and a pair $(i \in A, v \in V)$ is observed only when some planned solution actually assigns target $v$ to agent $i$.
To generate diverse target assignments, \planner adds noise of scale $\sigma$ to the current estimates and computes a target assignment on the result.
Concretely, we draw a matrix $E$ with i.i.d.\ $\mathcal{U}(0,1)$ entries and compute a target assignment on $\CostMatrix + \sigma E$ (\cref{algo:framework:noise}).
This results in diverse target assignments, as the noise changes the relative order of the entries.
The resulting assignment fixes the target of every agent and thus yields a MAPF instance.
This whole process---adding the noise, computing the assignment, and planning the MAPF instance---involves no interaction between repetitions, and thus \planner runs $P$ of them in parallel~(\cref{algo:framework:parallel}).
Our implementation uses $\sigma=5$ and $P=8$ by default. 

\paragraph{Cost Feedback (\cref{algo:framework:update}).}
In each iteration, the travel losses observed in a solution are reflected in $\CostMatrix$.
Let $c_i(\solutions_k)$ be the travel loss of agent $i$ with a target assignment $\assignment_k$.
We blend each observation $c_i(\solutions_k)$ into the corresponding entry $\CostMatrix[i][\assignment_k(i)]$ through a moving average.
When several candidates of the iteration assign the same pair, their observations are averaged first.
Concretely, let $K_{i, t} = \{k \mid \assignment_k(i) = t\}$ collect the candidates of the round that assigned target $t$ to agent $i$.
For every pair with $K_{i, t} \neq \emptyset$, we update
\begin{equation}
\CostMatrix[i][t] \leftarrow (1-\alpha)\, \CostMatrix[i][t]
+ \alpha \cdot \frac{1}{|K_{i,t}|} \sum_{k \in K_{i,t}} c_i(\solutions_k),
\label{eq:update}
\end{equation}
where $\alpha$ controls how strongly each new observation overrides the current estimate.
Our implementation sets $\alpha = 0.2$.
Each iteration also keeps the better of its $P$ solutions and the best solution so far (\crefrange{algo:framework:argmin}{algo:framework:track}).
The output is this best assignment with its planned paths, not the assignment implied by the final cost matrix.

\paragraph{Final Path Optimization (\cref{algo:framework:path-opt}).}
After the search loop terminates, the framework can optionally re-plan the paths for the best assignment found.
The search for target assignment favors a fast MAPF solver, because it must evaluate many candidates as quickly as possible.
The resulting paths are, however, typically suboptimal.
Applying a computationally heavier solver to the best assignment can therefore further improve the overall solution quality.
In our implementation, vanilla LaCAM~\cite{okumura2023lacam} evaluates the candidates during iterations, and LaCAM3~\cite{okumura2024lacamstar} optimizes the final paths.

\newlength{\evomatrixh}

\newcommand{\evocol}[4]{%
  \begin{minipage}[b]{0.178\textwidth}
    \centering
    iteration #1\\
    \mbox{improvement #2\,\%}\\[3pt]
    \includegraphics[width=\linewidth]{#3}\\[4pt]
    \includegraphics[width=\linewidth]{#4}%
  \end{minipage}}

\begin{figure*}[t!]
  \centering
  \setlength{\evomatrixh}{0.174\textwidth}
  \rotatebox{90}{\makebox[\evomatrixh][c]{agents}}%
  \evocol{0}{0.0}{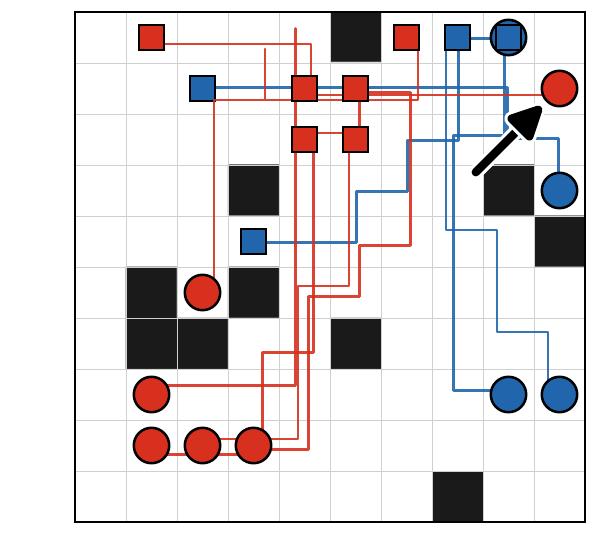}{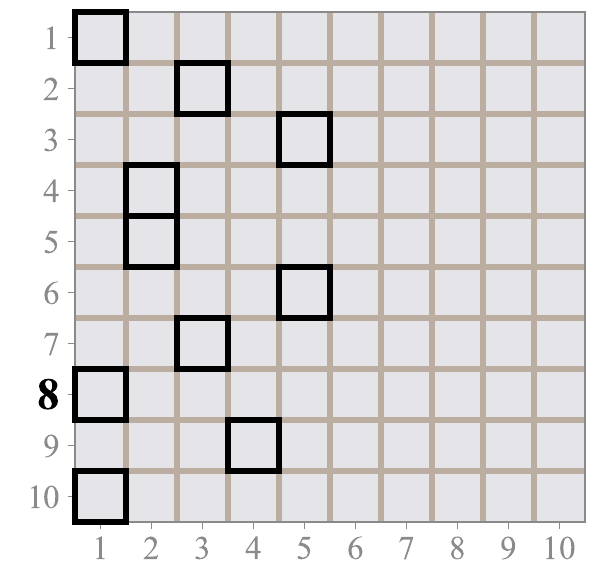}\hfill
  \evocol{2}{14.3}{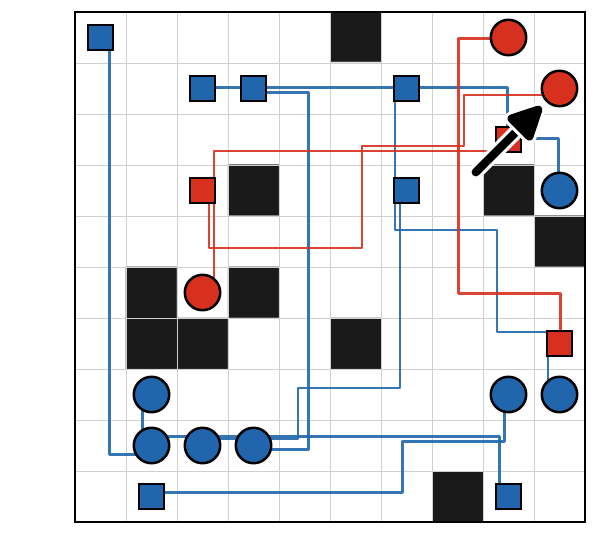}{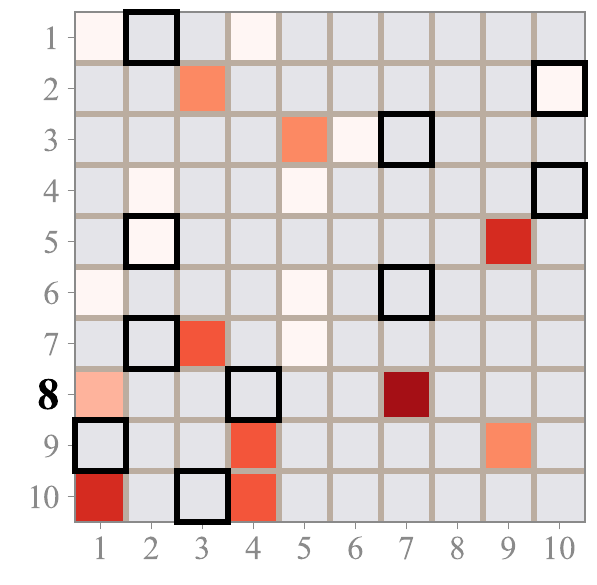}\hfill
  \evocol{5}{17.0}{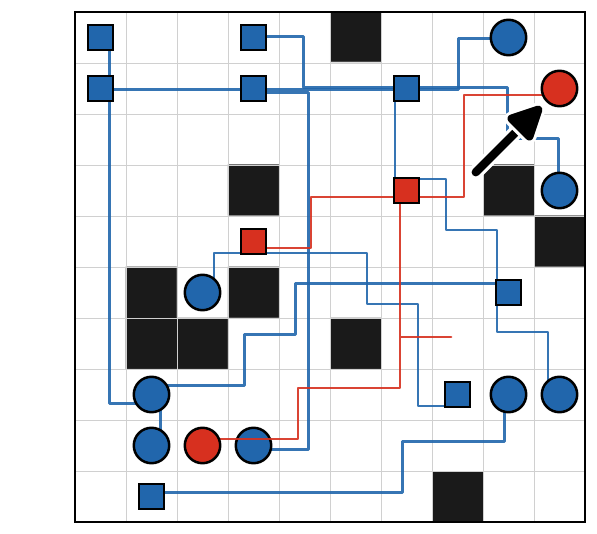}{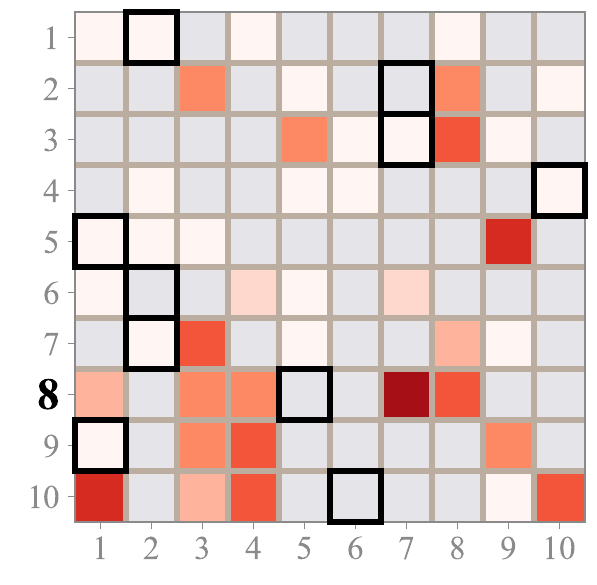}\hfill
  \evocol{12}{18.8}{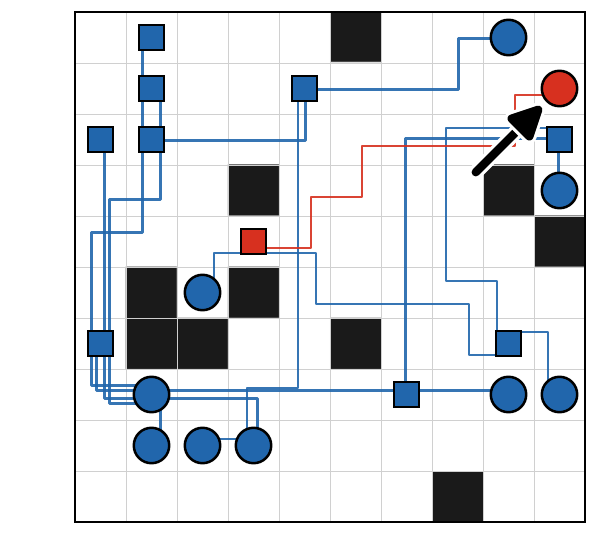}{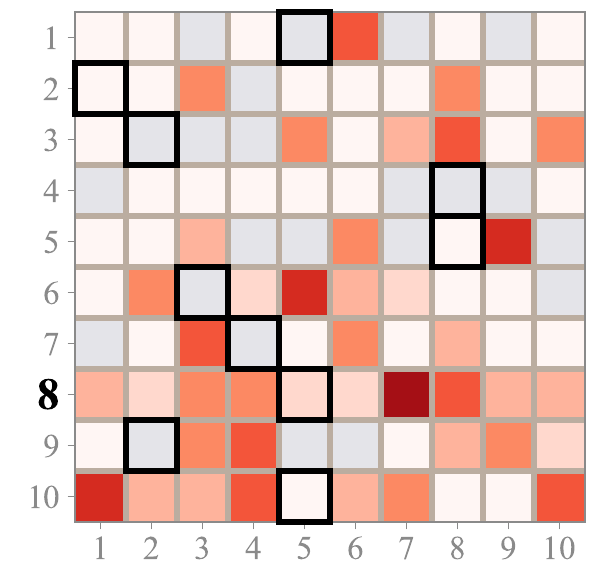}\hfill
  \evocol{73}{19.6}{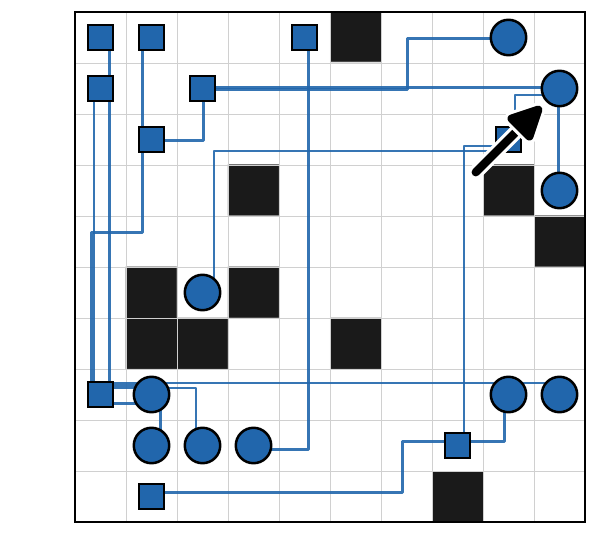}{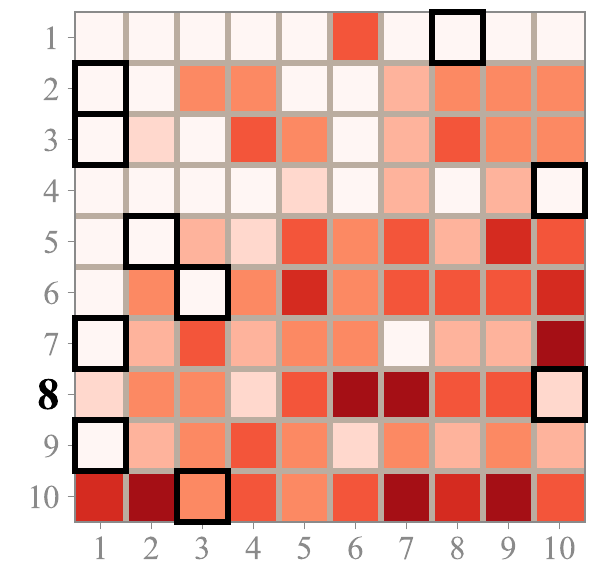}\hfill
  \begin{minipage}[b]{0.05\textwidth}
    \centering
    \includegraphics[height=\evomatrixh]{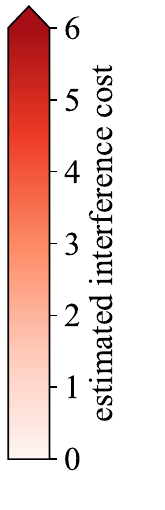}%
  \end{minipage}
  targets
  \caption{
  Evolution of the cost matrix on a 10$\times$10 grid with 10 agents.
  Each agent is given 10 candidate targets, all at the same shortest-path distance.
  Each column shows one iteration: the map (top) draws the paths at that iteration, with circles as agents and squares as their assigned
  targets.
  \textcolor{red}{Red paths contain delays} over the shortest-path distance (waits and detours), while \textcolor{blue}{blue paths do not}.
  The arrow marks agent~8, discussed in the main text.
  The cost matrix (bottom) shows the estimated interference cost ($\CostMatrix[i][v] - \dist(s_i, v)$), with rows as agents and columns as candidate targets. 
  Black frames mark the targets assigned at that iteration. 
  Unobserved entries remain gray.
  }
  \label{fig:costmatrix}
\end{figure*}
\subsection{Running Example}
\Cref{fig:costmatrix} shows how the cost matrix is calibrated over iterations and its effect on the solution.
The instance follows the \scenEqdist scenario used in our experiments (\cref{sec:setup}), where every agent receives its candidate targets at the same shortest-path distance from its start.
Under distance-based estimates, the choice of targets thus makes no difference.
In practice, however, interference between agents arises as in iteration~0.
As the bottom heatmaps show, \planner calibrates the travel cost while trying diverse target assignments.
For example, looking at agent~8, marked by the arrow, we see that its assigned target changes from iteration to iteration, and each attempt records a new estimate in its row of the matrix.
Between iterations 5 and 12, the assignment of agent~8 stays unchanged, but the assignments of the other agents change and their delays are resolved.
Finally, at iteration~73, the target assignment selects for agent~8 one of the smallest estimates in its row, and its path is free of delays for the first time.
The same holds for every other agent, and \planner reaches an optimal solution where the path of every agent matches its shortest-path distance.

\section{Evaluation}
This section empirically evaluates \planner.
The experiments consist of four parts:
\emph{(i)}~assessing the optimality on small instances;
\emph{(ii)}~assessing its solution quality on medium-scale instances,
\emph{(iii)}~evaluating its scalability with up to $10{,}000$ agents; and
\emph{(iv)}~analyzing the effect of each component.

\subsection{Setup}
\label{sec:setup}
Experiments were conducted on a Mac Studio with M1 Ultra (20 cores) and \SI{64}{\giga\byte} RAM.
We benchmark on four-connected grid maps from the MAPF benchmark~\cite{stern2019def}, containing diverse environments with varying sizes and structural characteristics.
On each map and each number of agents, we generate TAPF instances with a target matrix $\mathcal{T}$ constructed under one of the following two scenarios.
\begin{itemize}
    \item \scenEqdist: Each agent receives ten targets sampled uniformly from the vertices at a fixed shortest-path distance $d$ from its start, where $d$ is the median pairwise shortest-path distance of the map (e.g., $d=43$ on \mapname{random-64-64-20}).
    The values of $d$ for all maps are listed in the appendix.
    \item \scenHotspot: Following~\cite{kumagai2026alternating}, the targets of all agents are concentrated in a single densely packed region, the set of vertices within radius $r$ of a fixed center (e.g., $r = 35$ at $n = 400$ on \mapname{random-64-64-20}); more details are in the appendix.
    Within this hotspot, each agent receives ten targets, and the target sets of different agents overlap by about $80\%$.
\end{itemize}
The two scenarios stress different aspects of target assignment.
While \scenEqdist directly measures the importance of the target assignment as all targets of an agent are at the same shortest-path distance from its start, \scenHotspot poses more challenging instances with denser setups. 
A careful choice of targets within the hotspot is desired to steer agents away from the crowded spots.

As a solution metric, we report $\mathrm{flowtime} / \mathrm{LB}$, where $\mathrm{LB}$ is the value of an optimal linear assignment on the single-agent shortest-path distances.
$\mathrm{LB}$ is a lower bound on the achievable flowtime, and the reported ratio is at least $1.0$.

We compare the following representative TAPF methods:
\begin{itemize}
\item \textbf{ITA-ECBS}~\cite{tang2024ita-ecbs} is a bounded-suboptimal TAPF solver.
We tested two suboptimality factors: $w=1.0$ and $w=1.1$. 
\item \textbf{DBS-Hungarian (D-H)}~\cite{kumagai2026alternating} is the state-of-the-art scalable solver, which iteratively improves the target assignment for small groups of agents.
\item \textbf{SSP-LaCAM3} follows the trivial approach: it computes one linear assignment on the single-agent shortest-path distances and spends the remaining budget on the anytime MAPF solver, LaCAM3~\cite{okumura2024lacamstar}.
\item \textbf{\planner} is our proposed algorithm.
Unless otherwise noted, we set $\alpha = 0.2$, $\sigma = 5$, and $P = 8$.
\end{itemize}
All methods except ITA-ECBS start from the same initial target assignment, and every method receives the same time budget of \SI{10}{\second} per instance.
Unless otherwise noted, each reported value is the mean over 30 runs for each map and each number of agents.

\subsection{Optimality Assessment}
\begin{figure}[t!]
    \centering
    \begin{minipage}[c]{0.50\columnwidth}
        \centering
        \begin{tikzpicture}
            \node[inner sep=0] (img)
                {\includegraphics[width=\linewidth]{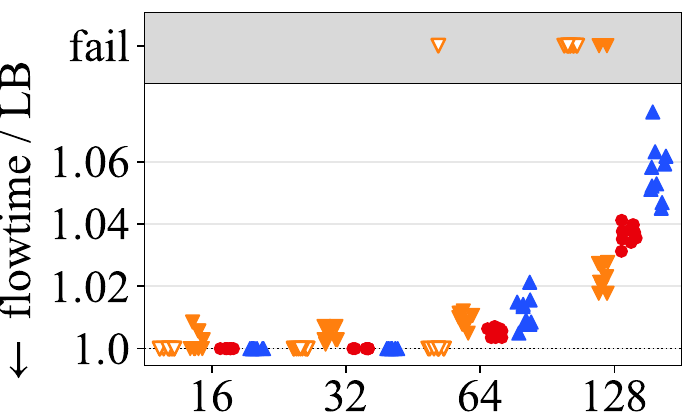}};
            \node[anchor=south, font=\small, yshift=1pt]
                at ([xshift=0.10\linewidth]img.north) {\scenEqdist};
        \end{tikzpicture}
    \end{minipage}\hfill
    \begin{minipage}[c]{0.50\columnwidth}
        \centering
        \begin{tikzpicture}
            \node[inner sep=0] (img)
                {\includegraphics[width=\linewidth]{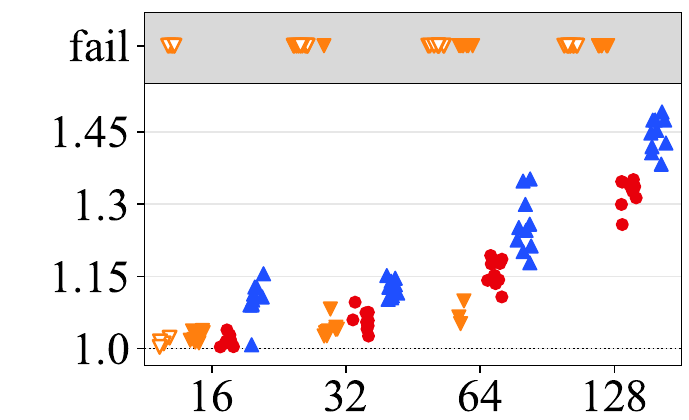}};
            \node[anchor=south, font=\small, yshift=1pt]
                at ([xshift=0.10\linewidth]img.north) {\scenHotspot};
        \end{tikzpicture}
    \end{minipage}
    {\centering\small agents}
    {\includegraphics[width=\linewidth]{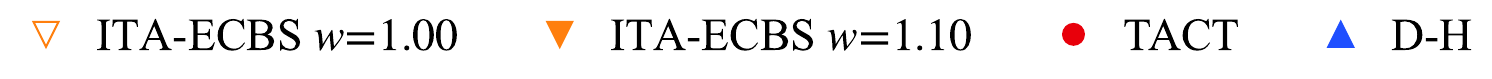}}
    \caption{
    Comparison with solvers on \emph{random-32-32-10} in \scenEqdist (left) and \scenHotspot (right), evaluated over 10 instances.
    Instances that ITA-ECBS fails to solve within the time limit are plotted in the gray band labeled ``fail.''
    }
    \label{fig:ita-ecbs-comparison}
\end{figure}
{
\setlength{\tabcolsep}{0pt}
\newcommand{\entryheadheight}{2.8em}
\newcommand{\mapplotsep}{0.0em}
\newcommand{\xlabellegendsep}{0.2em}
\newcommand{\mapentryshift}{0.16\linewidth}
\newcommand{\maplabel}[1]{%
  \ifthenelse{\equal{#1}{warehouse-10-20-10-2-1}}{%
    \mapname{warehouse}\\[-0.45em]
    \mapname{10-20-10-2-1}%
  }{%
    \ifthenelse{\equal{#1}{warehouse-20-40-10-2-1}}{%
      \mapname{warehouse}\\[-0.45em]
      \mapname{20-40-10-2-1}%
    }{%
      \ifthenelse{\equal{#1}{random-64-64-20}}{%
        \mapname{random}\\[-0.45em]
        \mapname{64-64-20}%
      }{%
        \ifthenelse{\equal{#1}{Paris_1_256}}{%
          \mapname{Paris}%
        }{%
          \mapname{#1}%
        }%
      }%
    }%
  }%
}
\newcommand{\mapentry}[2]{%
  \begin{minipage}[t]{0.12\linewidth}
    \centering
    \setlength{\baselineskip}{0.75\baselineskip}
    \scriptsize
    \hspace*{\mapentryshift}%
    \parbox[t][\entryheadheight][c]{\linewidth}{%
      \begin{minipage}[c]{0.28\linewidth}
        \centering
        \includegraphics[width=\linewidth]{fig/raw/maps/#1}
      \end{minipage}\hspace{-0.02\linewidth}%
      \begin{minipage}[c]{0.7\linewidth}
        \centering
        {\tiny\maplabel{#1}}\\[-0.2em]
        {\tiny\makebox[\linewidth][c]{(#2)}}%
      \end{minipage}%
    }%
  \end{minipage}%
}
\newcommand{\plotentry}[2]{%
  \begin{minipage}[t]{0.12\linewidth}
    \centering
    \includegraphics[width=\linewidth]{fig/raw/main_results/#1/#2_soc}
  \end{minipage}%
}
\newcommand{\rowlabel}[1]{%
  \rotatebox{90}{\shortstack{%
    \small #1\\[0.2em]
    \small $\leftarrow$ flowtime / LB}}%
}
\newcommand{\xlabel}{\small agents}
\begin{figure*}[t!]
  \centering
  \begin{tabular}{@{}c@{}cccccccc@{}}
    &
    \mapentry{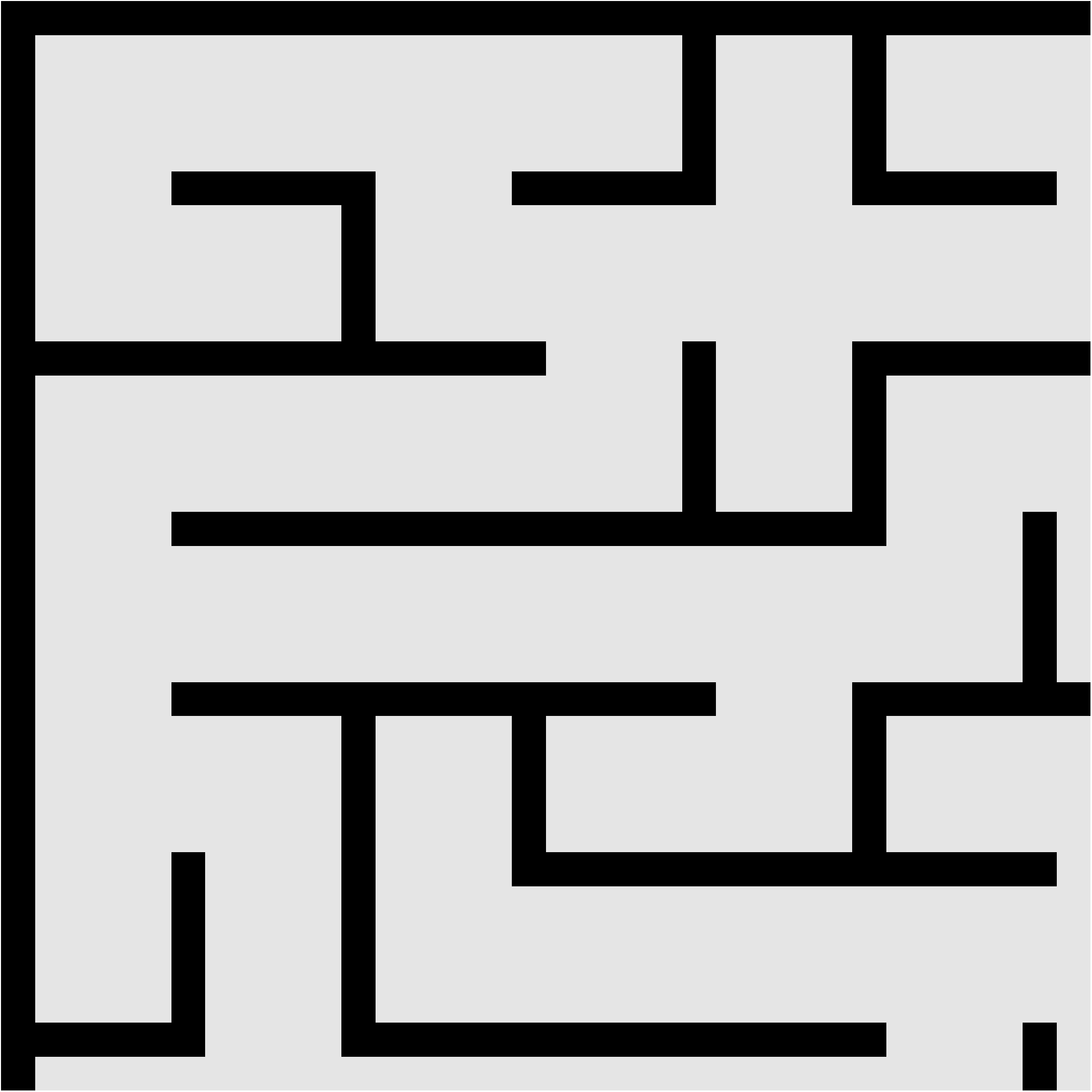}{32x32} &
    \mapentry{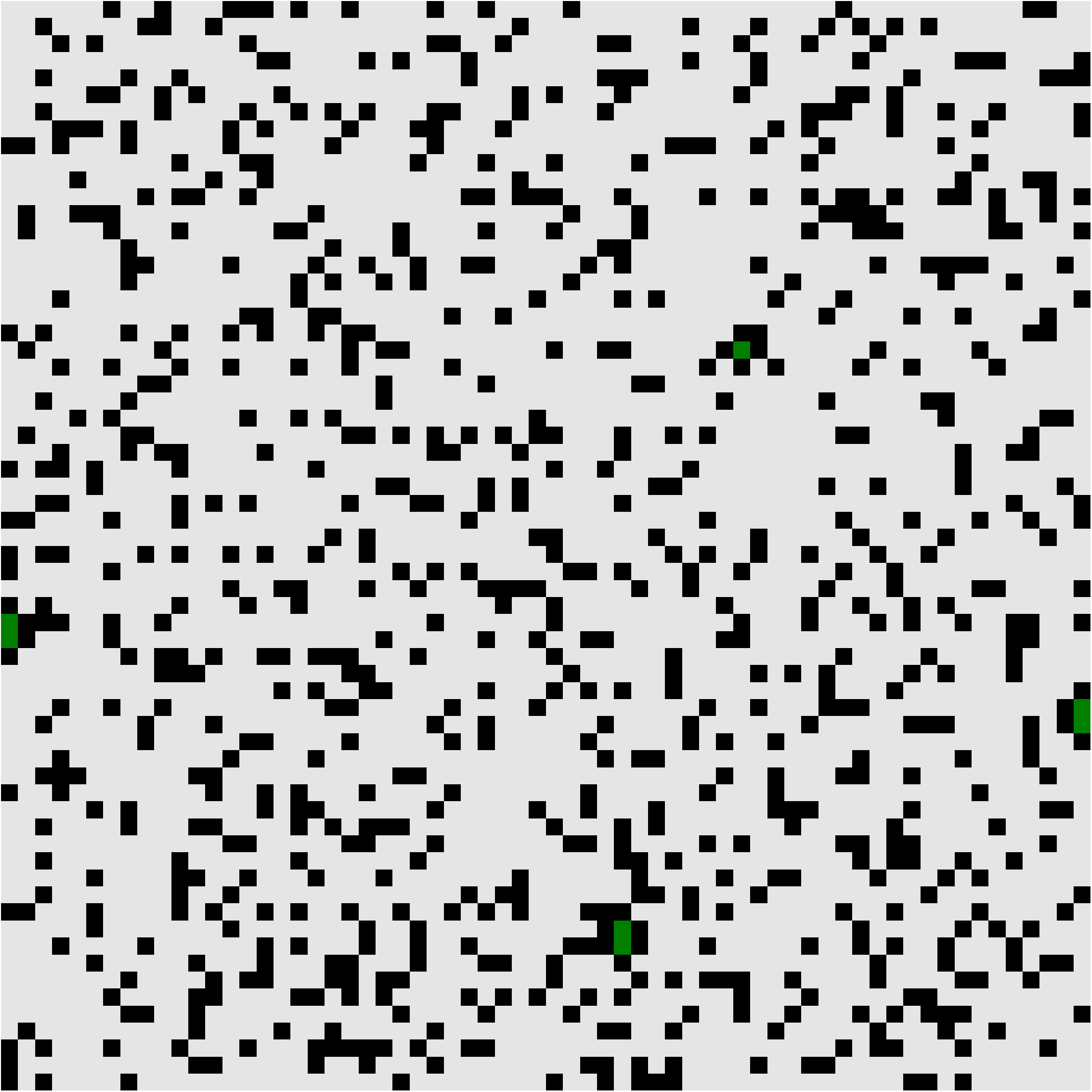}{64x64} &
    \mapentry{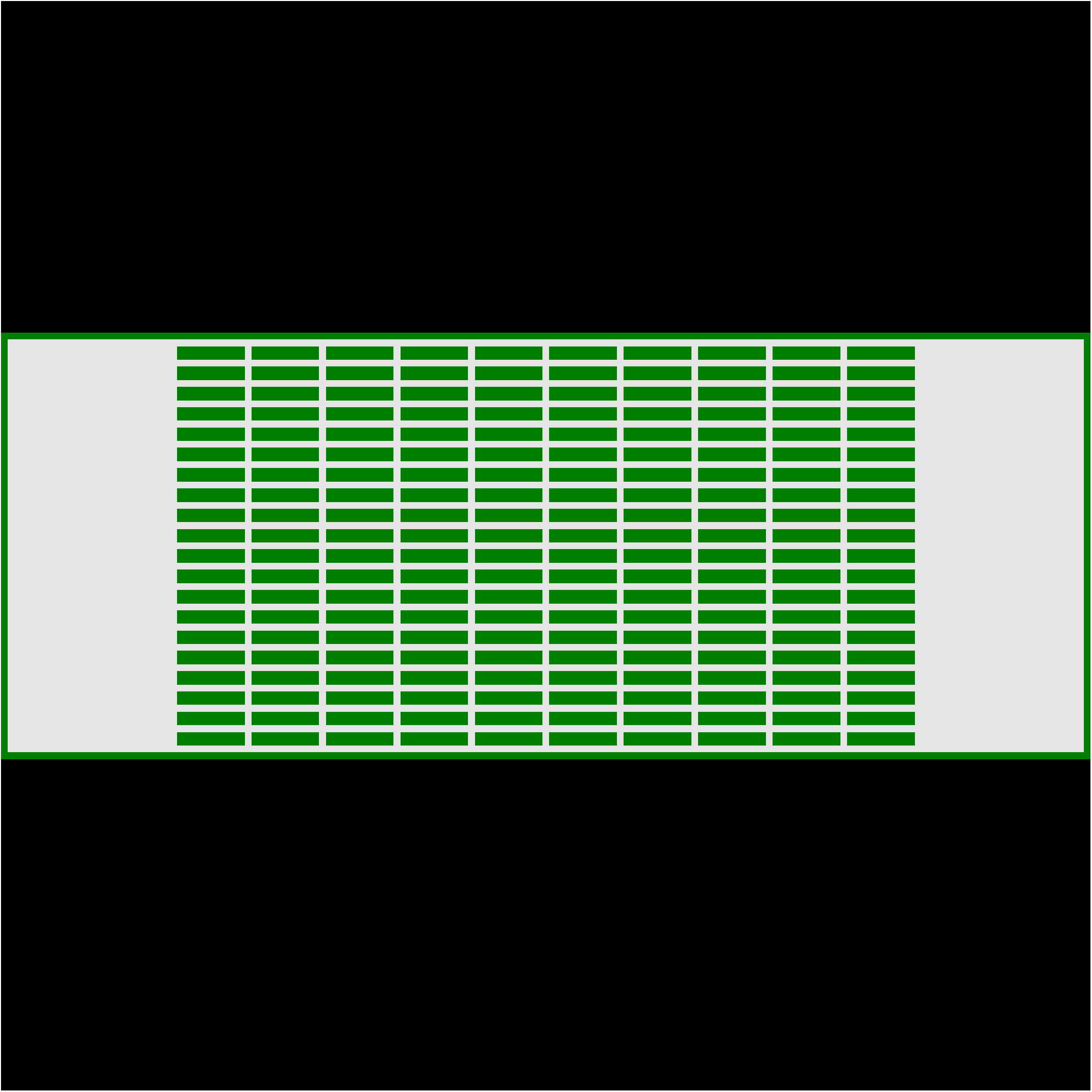}{161x63} &
    \mapentry{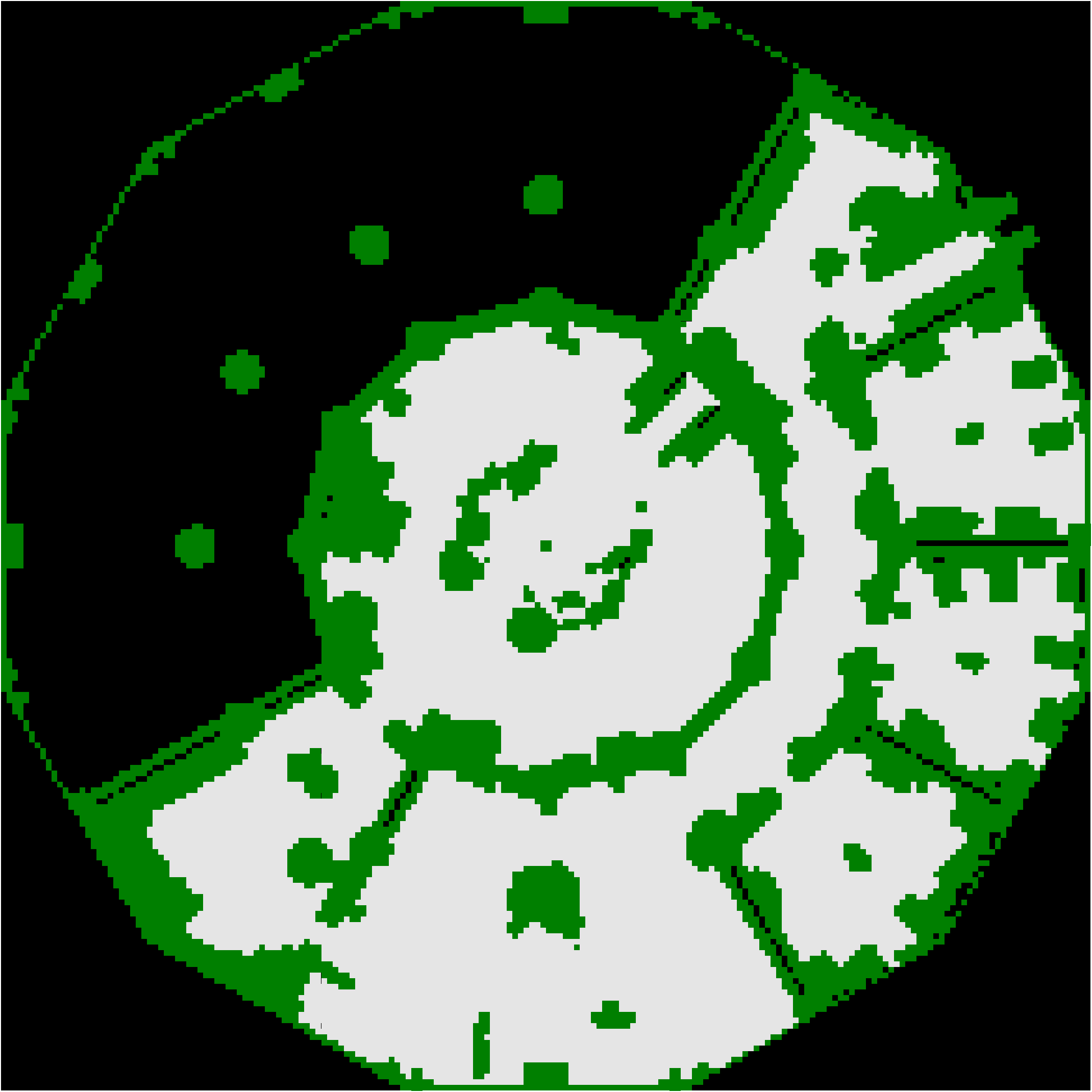}{194x194} &
    \mapentry{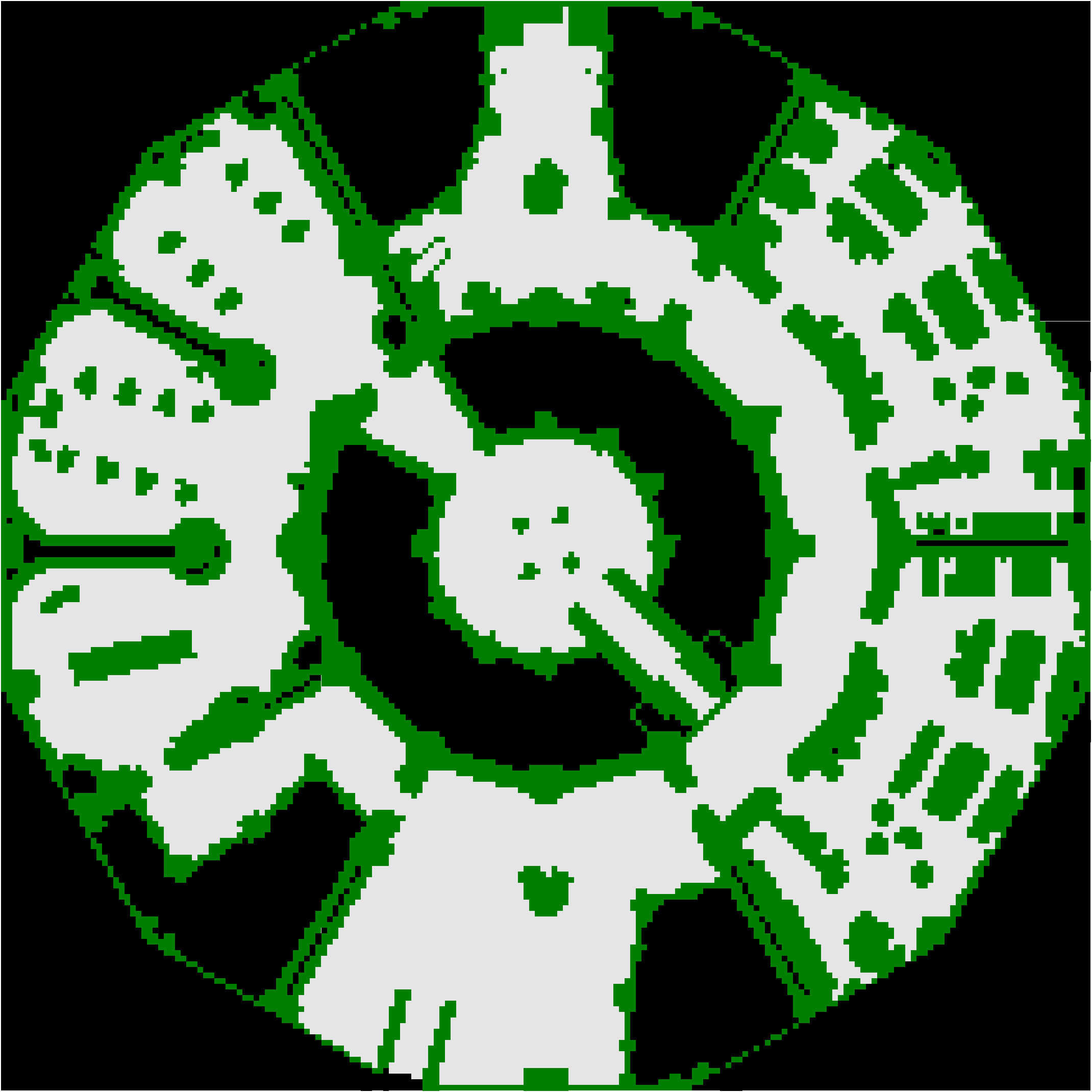}{194x194} &
    \mapentry{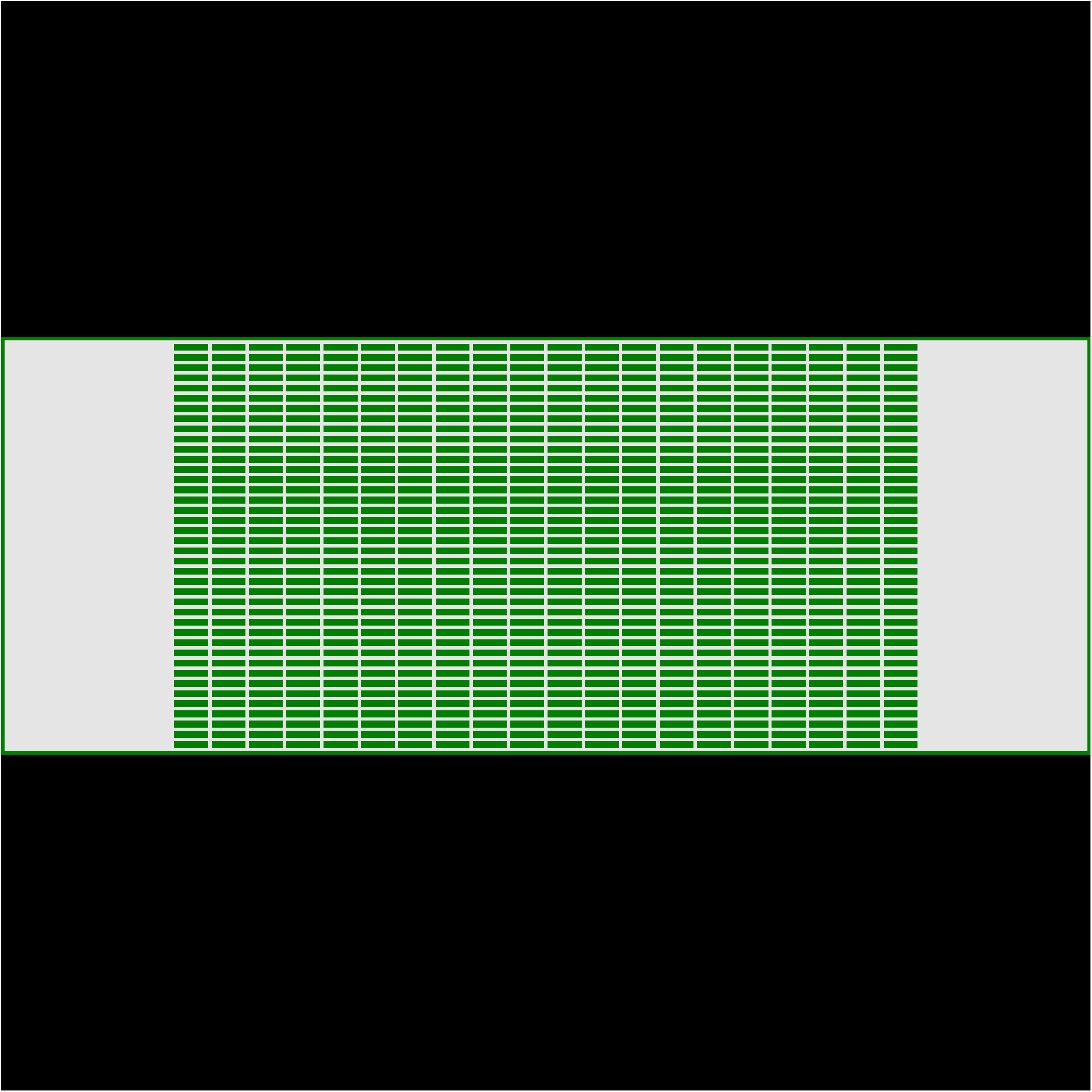}{321x123} &
    \mapentry{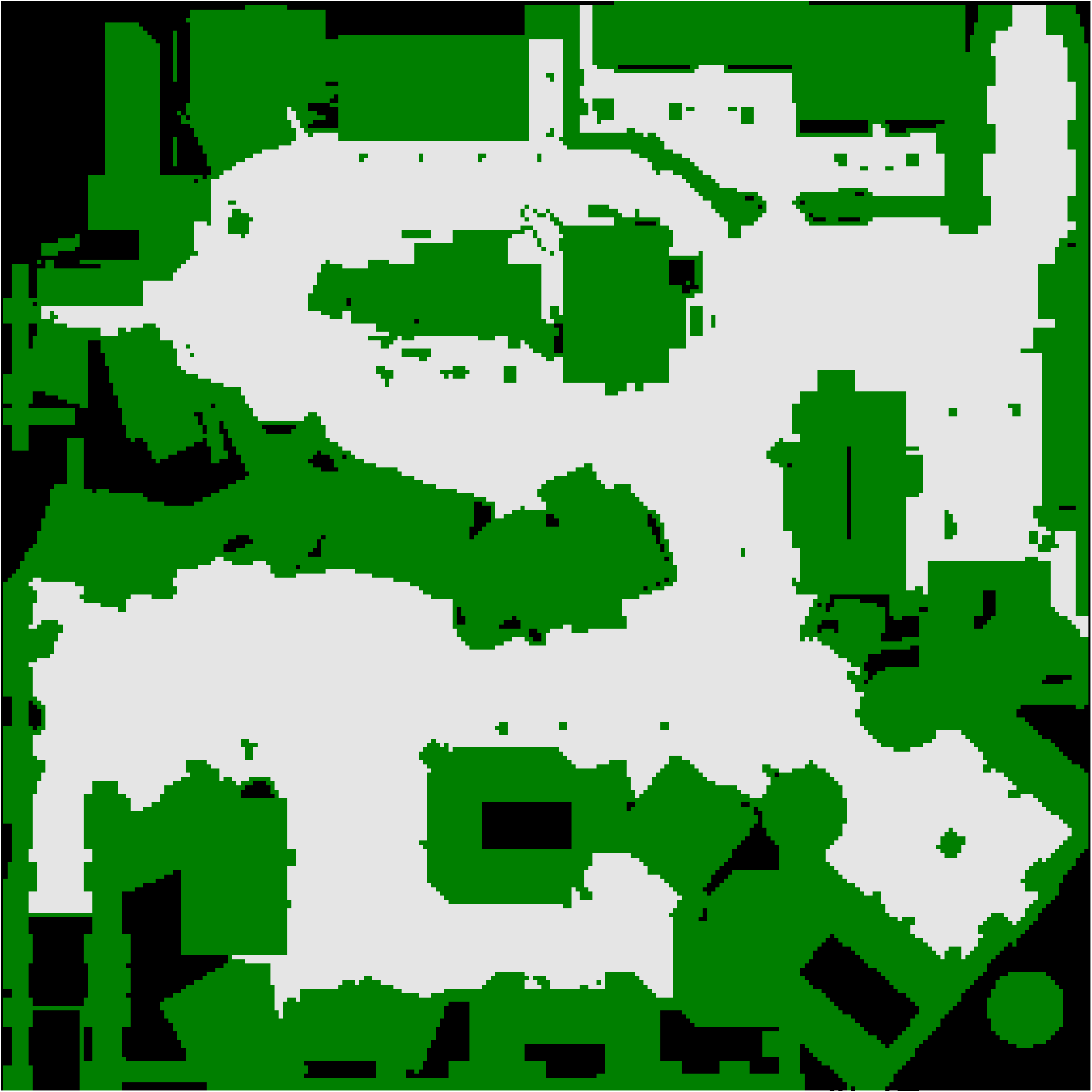}{256x257} &
    \mapentry{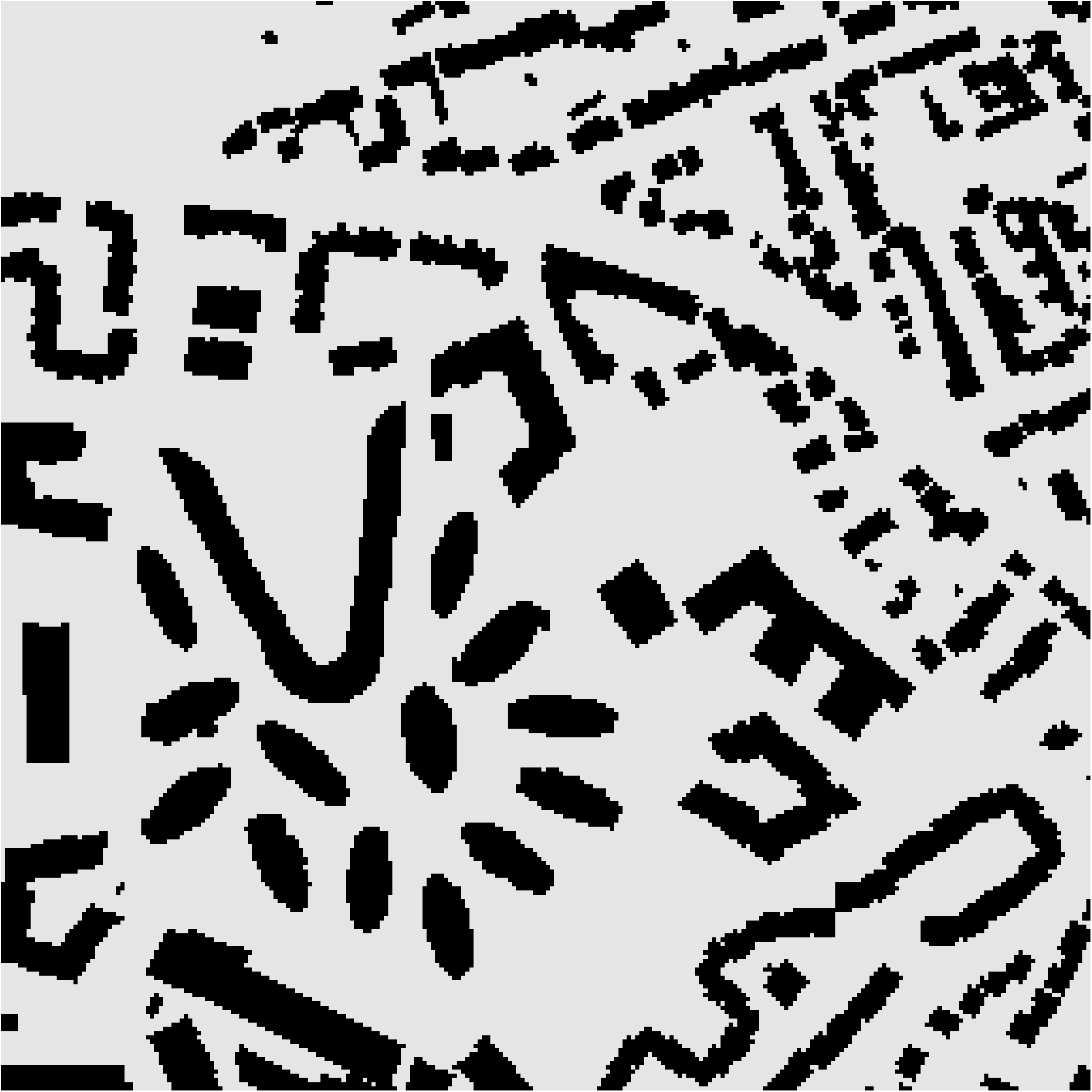}{256x256}
    \\[0.8em]
    \noalign{\vskip\mapplotsep}
    \rowlabel{\scenEqdist} &
    \plotentry{eqdist}{maze-32-32-4} &
    \plotentry{eqdist}{random-64-64-20} &
    \plotentry{eqdist}{warehouse-10-20-10-2-1} &
    \plotentry{eqdist}{ost003d} &
    \plotentry{eqdist}{lak303d} &
    \plotentry{eqdist}{warehouse-20-40-10-2-1} &
    \plotentry{eqdist}{den520d} &
    \plotentry{eqdist}{Paris_1_256}
    \\[0.3em]
    \rowlabel{\scenHotspot} &
    \plotentry{hotspot}{maze-32-32-4} &
    \plotentry{hotspot}{random-64-64-20} &
    \plotentry{hotspot}{warehouse-10-20-10-2-1} &
    \plotentry{hotspot}{ost003d} &
    \plotentry{hotspot}{lak303d} &
    \plotentry{hotspot}{warehouse-20-40-10-2-1} &
    \plotentry{hotspot}{den520d} &
    \plotentry{hotspot}{Paris_1_256}
    \\
    & \multicolumn{8}{c}{\xlabel}
    \\
    \noalign{\vskip\xlabellegendsep}
    & \multicolumn{8}{c}{\includegraphics[width=0.8\linewidth]{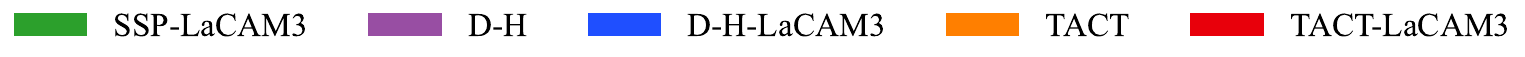}}
    \\
  \end{tabular}
  \caption{
  Solution quality on eight maps in \scenEqdist (top) and \scenHotspot (bottom) scenarios. 
  Lines show the mean scores, with the semi-transparent regions indicating a standard deviation.
  }
  \label{fig:result-main}
\end{figure*}
}
We first examine how close \planner comes to the optimum on small instances.
\Cref{fig:ita-ecbs-comparison} reports the results.
Under \scenEqdist, ITA-ECBS solves all but one instance up to $n=64$, and its optimal cost coincides with the lower bound on every solved instance.
\planner attains the optimum exactly up to $n=32$ and stays within $1\%$ at $n=64$.
At $n=128$, where the optimal reference no longer exists, the lower bound gap of \planner is still below $5\%$, while DBS-Hungarian grows to $6\%$ on average.
Under \scenHotspot, an optimal reference exists only at $n=16$, where ITA-ECBS with $w=1.0$ solves four instances.
On these instances, \planner stays near-optimal, within $0.5\%$ of the optimum on average.
As $n$ grows, ITA-ECBS fails to solve.
In contrast, \planner solves every instance, and it stays closer to the lower bound than DBS-Hungarian at every $n$. 

In summary, \planner finds (near-)optimal solutions wherever an optimality reference exists, and it remains close to the lower bound beyond the reach of ITA-ECBS.

\begin{table}[h!]
  \centering
  \small
  \setlength{\tabcolsep}{4pt}
  \renewcommand{\arraystretch}{0.92}
  \begin{tabular}{@{}rlrrrrrr@{}}
    \toprule
    $n$ & Solver & Imprv & Path & Total & Assign & LaCAM & Iter. \\
        &        & (\%)  & (\%) & (\%)  & [ms]   & [ms]  &       \\
    \midrule
    \multirow{2}{*}{200} & \planner-L3 & 14.7 & 2.4 & \textbf{16.7} & 0.21 & 9.7  & 628 \\
                         & D-H-L3      & 7.9  & 1.2 & 9.0           & 0.01 & 6.7  & 892 \\
    \midrule
    \multirow{2}{*}{400} & \planner-L3 & 12.2 & 4.3 & \textbf{16.0} & 0.36 & 17.1 & 358 \\
                         & D-H-L3      & 4.1  & 2.2 & 6.1           & 0.02 & 11.9 & 503 \\
    \midrule
    \multirow{2}{*}{600} & \planner-L3 & 9.4  & 6.9 & \textbf{15.6} & 0.52 & 25.5 & 238 \\
                         & D-H-L3      & 1.4  & 6.3 & 7.6           & 0.03 & 16.9 & 355 \\
    \midrule
    \multirow{2}{*}{800} & \planner-L3 & 6.3  & 9.2 & \textbf{14.9} & 0.69 & 29.8 & 173 \\
                         & D-H-L3      & 0.8  & 11.6 & 12.3         & 0.04 & 24.7 & 271 \\
    \bottomrule
  \end{tabular}
\caption{Detailed comparison of \planner-LaCAM3 (\planner-L3) and
    D-H-LaCAM3 (D-H-L3) on \mapname{random-64-64-20} \scenHotspot.
    \textit{Imprv} and \textit{Path} denote the mean flowtime improvement (\%) over the initial solution and by the final path optimization, respectively.
    \textit{Assign} and \textit{LaCAM} denote the mean runtime per iteration of the assignment step and of the LaCAM evaluation, respectively.
    \textit{Iter.} is the mean number of iterations for target reassignment.}
  \label{tab:detail-random64}
\end{table}

\subsection{Solution Quality Assessment}
We next assess the solution quality of \planner on medium-scale instances.
The final solution quality depends not only on the target assignment but also on the paths planned for it.
We therefore add two variants with path optimization: \textbf{\planner-LaCAM3} and \textbf{D-H-LaCAM3} spend the first \SI{6}{\second} of the budget on revising the target assignment and the remaining \SI{4}{\second} on the same path optimization as SSP-LaCAM3.

As \cref{fig:result-main} shows, \planner outperforms D-H on every map and every number of agents, demonstrating that the calibration succeeds in selecting good target assignments.
Taking SSP-LaCAM3 as a reference point, \planner alone outperforms it in some cases of \scenEqdist, and on most maps under \scenHotspot, meaning that in these cases, selecting good targets is more important for the solution quality than improving the paths.
Turning to the variants with path improvement, \planner-LaCAM3 and D-H-LaCAM3 outperform SSP-LaCAM3, which also confirms that the target assignment matters for the final solution quality.
Finally, \planner-LaCAM3 achieves the best solution quality in nearly all combinations.
For example, on \mapname{warehouse-10-20-10-2-1} at $n=800$ under \scenEqdist, \planner-LaCAM3 attains a ratio of $1.10$ while D-H-LaCAM3 and SSP-LaCAM3 remain at $2.02$ and $2.44$.
Combining good target assignment with path improvement therefore yields the best final solution quality.

\begin{figure}[t!]
    \centering
    \begin{minipage}[c]{0.03\columnwidth}
        \centering
        \rotatebox{90}{\small $\leftarrow$ flowtime / LB}
    \end{minipage}\hfill
    \begin{minipage}[c]{0.475\columnwidth}
        \centering
        {\small \mapname{random-64-64-20}, $n=400$}\\[1pt]
        \includegraphics[width=\linewidth]{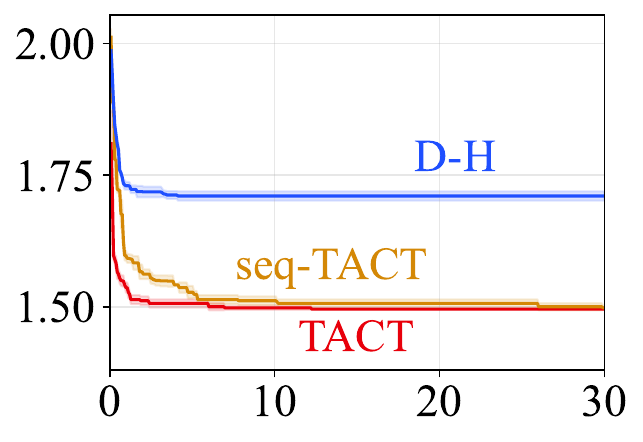}
    \end{minipage}\hfill
    \begin{minipage}[c]{0.475\columnwidth}
        \centering
        {\small \mapname{Paris\_1\_256}, $n=800$}\\[1pt]
        \includegraphics[width=\linewidth]{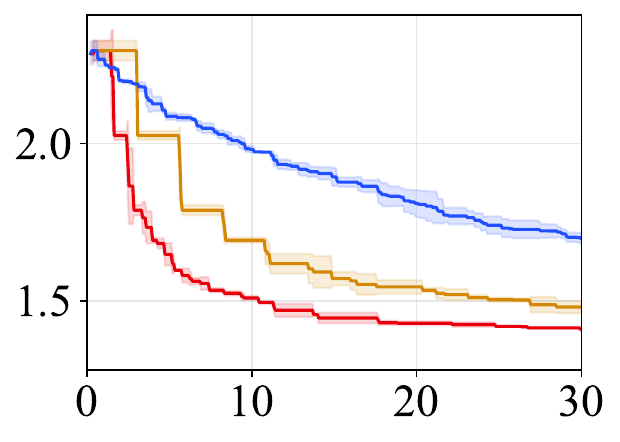}
    \end{minipage}\\[2pt]
    {\small runtime [s]}
    \caption{
    Anytime behavior of \planner, \planner with sequential evaluation~(seq-\planner), and DBS-Hungarian (D-H) for a \SI{30}{\second} budget in \scenHotspot.
    }
    \label{fig:convergence}
\end{figure}
Next, we look at the details of \planner-LaCAM3.
\Cref{tab:detail-random64} decomposes the improvement from the initial solution on \mapname{random-64-64-20} in \scenHotspot into the contribution of the target assignment search (\emph{Imprv}) and that of the final path optimization (\emph{Path}).
The \emph{Imprv} column shows that \planner gains more from revising the target assignment than DBS-Hungarian at every number of agents, e.g., $6.3\%$ versus $0.8\%$ at $n=800$.
\planner also reaches these gains in fewer iterations, because it accumulates the observed travel losses into the cost matrix and a single iteration can revise the assignment of all agents at once, whereas an iteration of DBS-Hungarian consumes its observation in one local reassignment within a small subgroup.
With the gains from both parts, \planner attains the largest total improvement in every setting.
\Cref{tab:detail-random64} also shows that LaCAM dominates the runtime of one iteration at every number of agents, so target assignment does not become the bottleneck as the number of agents grows.

\Cref{fig:convergence} examines the anytime behaviors.
\planner reduces the flowtime sharply within the first few seconds and keeps a clear margin over DBS-Hungarian.
To isolate the effect of parallel execution, we also run \planner in a sequential paradigm, labeled \emph{seq-\planner}, which evaluates the $P=8$ candidates one by one instead of in parallel.
The sequential variant converges slightly more slowly than \planner, but it still improves the solution far faster than DBS-Hungarian.
The fast improvement of \planner thus comes not from parallel execution but from its feedback calibration design.

\subsection{Scalability Test}
\begin{figure}[t!]
    \begin{minipage}[t]{0.414\columnwidth}
        \centering
        \includegraphics[width=\linewidth]{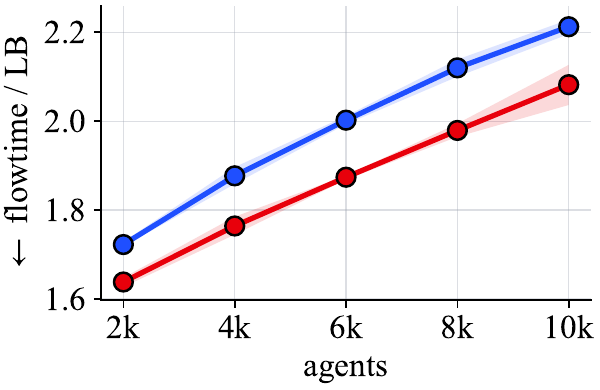}
        \includegraphics[height=0.16in]{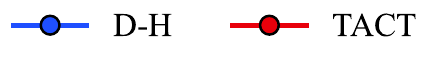}
    \end{minipage}\hfill
    \begin{minipage}[t]{0.566\columnwidth}
        \centering
        \includegraphics[width=\linewidth]{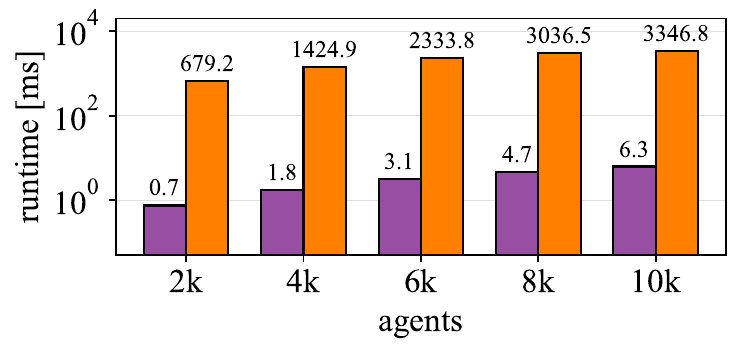}
        \includegraphics[height=0.16in]{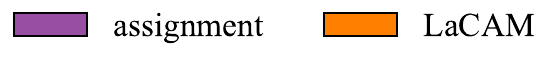}
    \end{minipage}
    \caption{
    Scalability on \mapname{warehouse-20-40-10-2-2} in \scenHotspot with $n=2{,}000$ to $10{,}000$ under a fixed \SI{5}{\minute} budget, and runtime of \planner, broken down into the target assignment and the LaCAM (log scale).
    }
    \label{fig:scalability}
\end{figure}
Next, \cref{fig:scalability} evaluates the scalability of \planner.
Given the problem scale, we set the time budget to \SI{5}{\minute}.
Even at this scale, one iteration takes around \SI{3}{\second}, and \planner still completes dozens of target reassignments and improves the solution quality.
These results demonstrate that \planner solves TAPF practically even at massive scale.

\subsection{Ablation Study}
\label{sec:design}
The evaluation so far uses the following hyperparameters:
\begin{itemize}
\item The cost feedback learning rate $\alpha$ (default $0.2$) controls how strongly each observation moves the estimate.
\item The noise scale $\sigma$ (default $5$) controls how far the candidates scatter around the current cost matrix estimates.
\item $P$ (default $8$) is the number of MAPF feedback runs in parallel per iteration.
\end{itemize}
We now vary each parameter and examine how it shapes the behavior of \planner.

\paragraph{Feedback is necessary.}
\begin{figure}
    \centering
    \begin{minipage}[c]{0.49\columnwidth}
        \centering
        {\small \mapname{ost003d}}\\[1pt]
        \includegraphics[width=\linewidth]{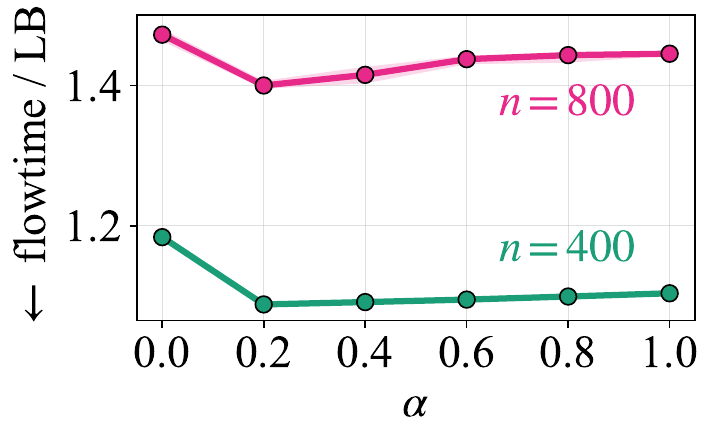}
    \end{minipage}\hfill
    \begin{minipage}[c]{0.49\columnwidth}
        \centering
        {\small \mapname{random-64-64-20}}\\[1pt]
        \includegraphics[width=\linewidth]{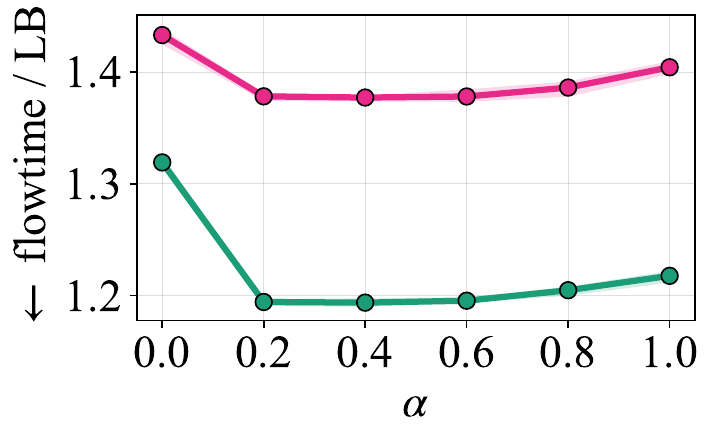}
    \end{minipage}
    \caption{
    Effect of the feedback rate $\alpha$, with the remaining parameters fixed to $\sigma = 5$ and $P = 8$.
    }
    \label{fig:alpha}
\end{figure}
\Cref{fig:alpha} sweeps $\alpha$ on two maps.
$\alpha = 0$ is clearly the worst in every panel.
This setting discards the observations entirely, so every iteration reduces to adding noise to the shortest-path distances.
The gap to the positive rates therefore shows that calibrating the estimates with observed travel costs, not the noise alone, is what drives the search toward good target assignments.
Among the positive rates, the differences are small, and we adopt $\alpha = 0.2$, where the curves reach their minimum or come close to it.

\paragraph{Diversity of Target Assignment Candidates.}
\begin{figure}[t!]
    \centering
    \mapname{random-64-64-20, $n=800$}\\[1pt]
    \begin{minipage}[c]{0.49\columnwidth}
        \centering
        \includegraphics[width=\linewidth]{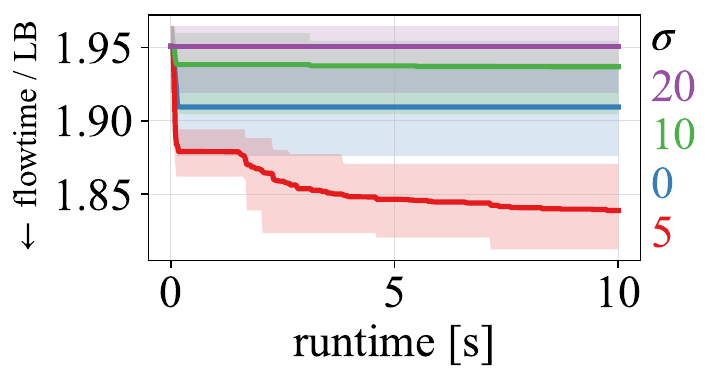}
    \end{minipage}\hfill
    \begin{minipage}[c]{0.49\columnwidth}
        \centering
        \includegraphics[width=\linewidth]{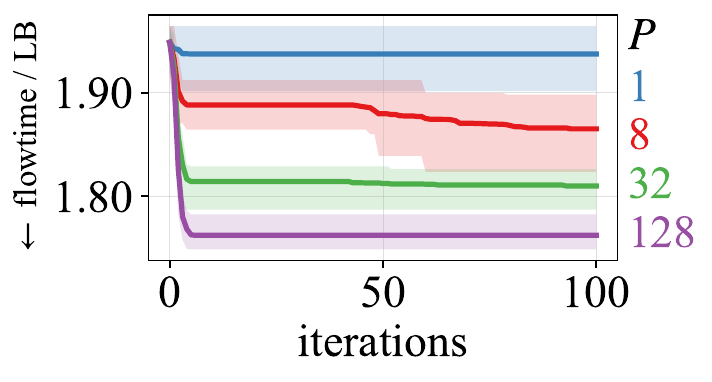}
    \end{minipage}
    \mapname{maze-32-32-4, $n=200$}\\[1pt]
    \begin{minipage}[c]{0.49\columnwidth}
        \centering
        \includegraphics[width=\linewidth]{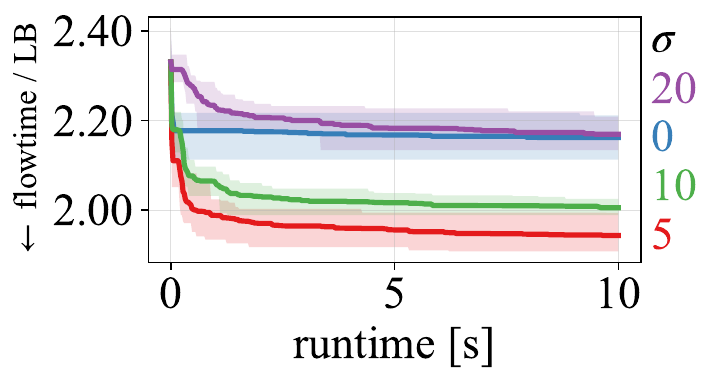}
    \end{minipage}\hfill
    \begin{minipage}[c]{0.49\columnwidth}
        \centering
        \includegraphics[width=\linewidth]{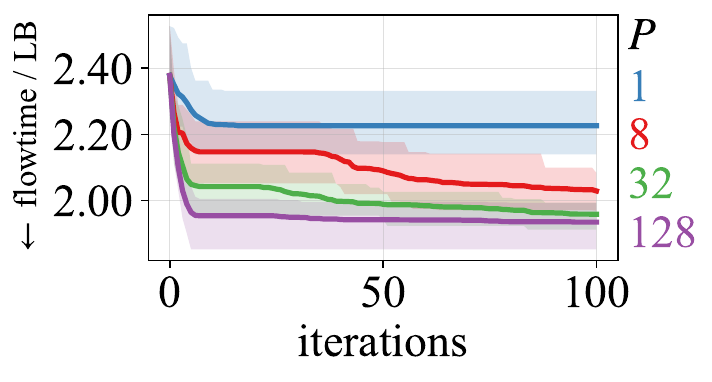}
    \end{minipage}
    \caption{
    Effect of the noise scale $\sigma$ (left column) and the number of
    parallel candidates $P$ (right column), with feedback rate $\alpha = 0.2$.
    The left column shows the best flowtime over the lower bound against
    runtime with $P = 8$ fixed. The right column shows it against
    iterations with $\sigma = 5$ fixed.
    }
    \label{fig:sigma_parallel}
\end{figure}
The left column of \cref{fig:sigma_parallel} shows that a large $\sigma$ degrades the solution quality, because the noise dominates the estimates and the candidates become almost random.
At the other end, $\sigma=0$ collapses the $P$ candidates of an iteration into a single target assignment.
\planner still improves the solution under $\sigma=0$, because the feedback keeps revising the cost matrix and thereby changes the assignment across iterations.
The trace nevertheless shows the cost of this lost diversity, as both maps settle at a worse solution than with $\sigma=5$.
The right column shows that a larger $P$ reaches a good solution within fewer iterations by trying more candidates per iteration, at the price of wall-clock time since each iteration waits for the slowest of the $P$ planning runs.
Based on these observations, we adopt $\sigma = 5$ and $P = 8$.

\section{Conclusion and Discussion}
This paper presented \planner, which calibrates the travel costs through repeated path planning feedback with a fast suboptimal MAPF solver, collecting observations across varied target assignments.
Experiments reveal that, in large instances, \planner achieves the best solution quality in nearly all combinations of maps and numbers of agents.
The effectiveness of \planner comes from the interplay of its components.
Computing a target assignment is cheap enough to be tried many times within the time budget, and the added noise turns these trials into varied target assignments, each rearranging the targets of many agents rather than a small subset.
Every trial returns valuable information: the travel losses observed in actual path planning.
Then, \planner accumulates it in the cost matrix instead of discarding it.
In this way, \planner tries diverse target assignments a sufficient number of times and wastes none of the feedback, which explains its consistent advantage in solution quality.
\section*{Acknowledgement}
This study was partially supported by JSPS KAKENHI Grant Number 25K21289 and JST PRESTO (JPMJPR2513).
\bibliography{sty/ref-macro,ref}
\appendix
\section*{Appendix}
\suppressfloats[t]

\section{Other Experimental Scenarios}
The main text evaluates \planner under \scenHotspot and \scenEqdist.
This section covers two other scenarios from prior work and explains why they are not adopted in the main text.
\begin{figure}[t]
    \centering
    \includegraphics[width=\linewidth]{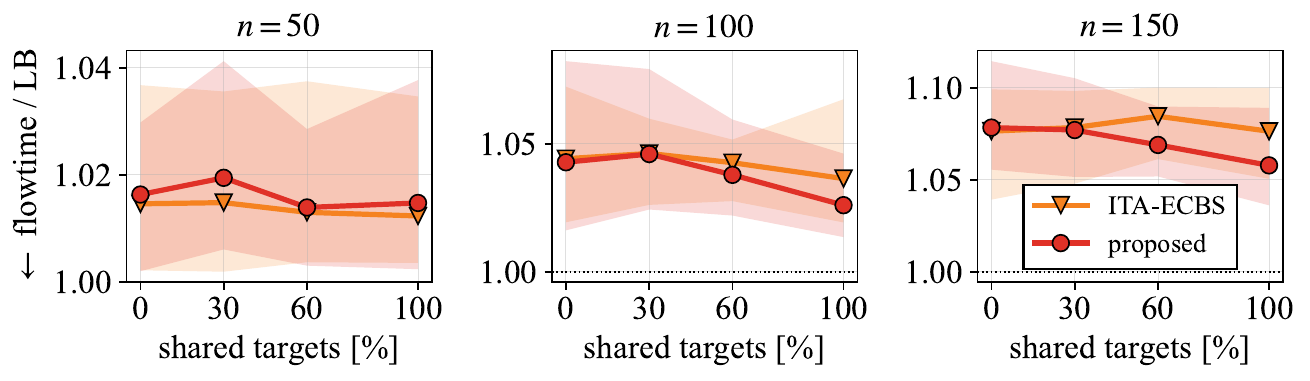}
    \caption{
    Comparison with ITA-ECBS ($w = 1.10$) in the shared-target setting
    of ITA-ECBS~\cite{tang2024ita-ecbs} on \mapname{random-32-32-10} with
    $K = 5$. Each panel reports the flowtime over the lower bound
    against the percentage of shared targets, as mean over 10 instances
    with a time limit of \SI{10}{\second} per solver. Shaded regions
    indicate the min--max range.
    }
    \label{fig:shared-ita-ecbs}
\end{figure}
\begin{figure}[t]
    \centering
    \begin{minipage}[c]{0.035\columnwidth}
        \centering
        \rotatebox[origin=c]{90}{\footnotesize $\leftarrow$ flowtime / LB}
    \end{minipage}\hfill
    \begin{minipage}[c]{0.31\columnwidth}
        \centering
        \includegraphics[width=\linewidth]{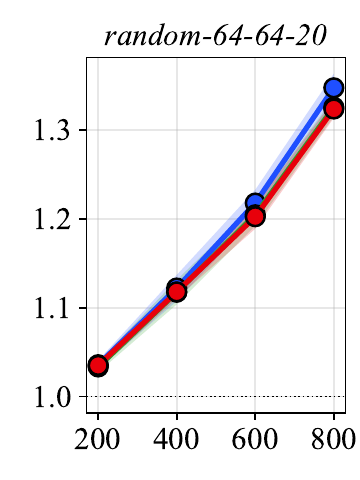}
    \end{minipage}\hfill
    \begin{minipage}[c]{0.31\columnwidth}
        \centering
        \includegraphics[width=\linewidth]{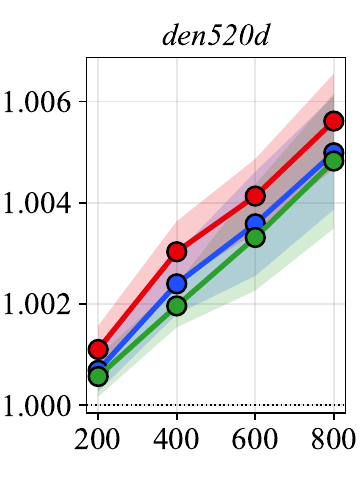}
    \end{minipage}\hfill
    \begin{minipage}[c]{0.31\columnwidth}
        \centering
        \includegraphics[width=\linewidth]{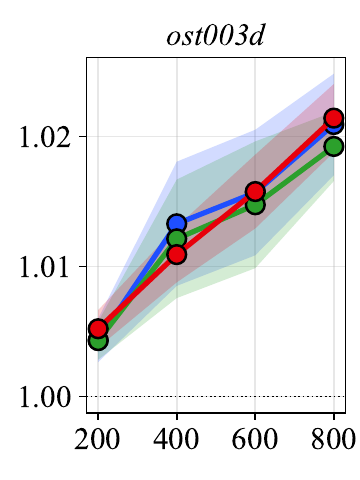}
    \end{minipage}\\[1pt]
    \hspace*{0.035\columnwidth}{\footnotesize agents}\\[3pt]
    \includegraphics[width=0.8\columnwidth]{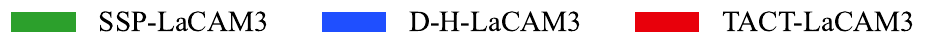}
    \caption{Comparison on fully random scenario.}
    \label{fig:random_main}
\end{figure}

\paragraph{Shared-Target Scenario.}
The first is the setting from the ITA-ECBS paper~\cite{tang2024ita-ecbs}.
It places starts and targets uniformly at random and controls the overlap of the target sets by a shared-target percentage: each agent receives a mix of targets shared among all agents and private targets that are disjoint across agents.
This design, however, does not extend to large numbers of agents, because the disjoint private targets consume vertices in proportion to the number of agents, so the target set size must shrink as the instance grows.
The original paper caps the target set size at the number of free cells divided by the maximum number of agents, and our reproduction with $K=5$ on \mapname{random-32-32-10} already uses 900 of the 922 free cells at $n=150$ with no shared targets.
Within the range the setting supports, we compare \planner with ITA-ECBS ($w = 1.1$).
As \cref{fig:shared-ita-ecbs} shows, both solvers stay within a few percent of the lower bound everywhere, and no percentage of shared targets separates them.
The only visible gap appears at $n=150$ with high shared percentages, where \planner is slightly ahead.

\paragraph{Fully Random Scenario.}
The second, following \cite{kumagai2026alternating}, samples the $K$ targets of each agent uniformly from the whole map.
As \cref{fig:random_main} shows, the compared solvers again end up close to one another. 

\section{Sigma Annealing}
\begin{figure}[t!]
    \centering
    {\small \mapname{random-64-64-20}, $n=400$}\\[1pt]
    \includegraphics[width=\columnwidth]{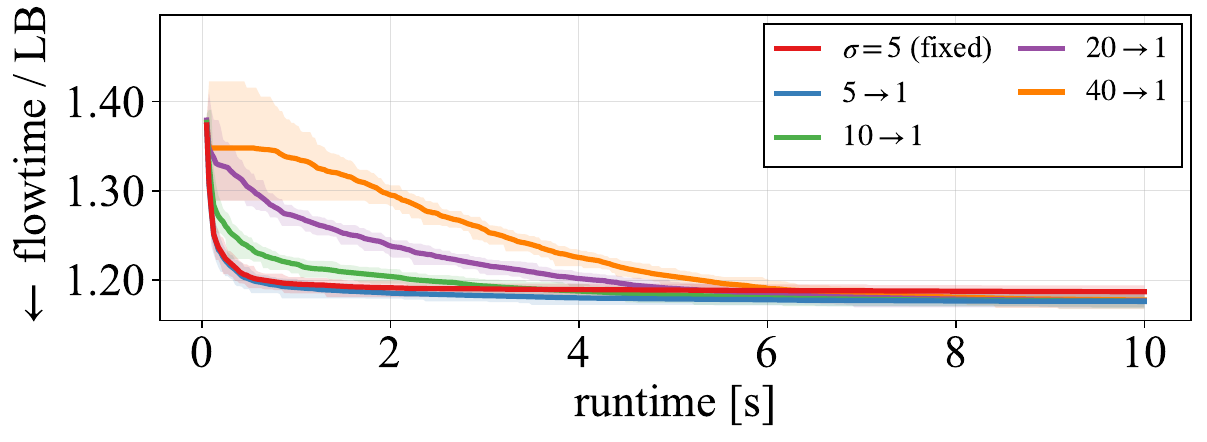}
    \caption{
    Annealing of the noise scale from $\sigma_0 \in
    \{5, 10, 20, 40\}$ down to $1$, against the constant $\sigma = 5$,
    in \scenHotspot.
    }
    \label{fig:annealing}
\end{figure}
Instead of keeping the noise scale fixed, one could anneal it.
\planner would then start with a large $\sigma$ that reassigns targets aggressively and end with a small $\sigma$ that only fine-tunes the assignment.
\Cref{fig:annealing} shows the convergence.
Every annealed variant indeed ends below the constant $\sigma = 5$, but the margin is below $1\%$ of flowtime.
The benefit of annealing is thus marginal, and we keep the constant $\sigma$ in the main configuration.

\section{Why \planner Returns Iteration Best}
\begin{figure}[t!]
    \centering
    \begin{minipage}[c]{0.49\columnwidth}
        \centering
        \begin{tikzpicture}
            \node[inner sep=0] (img)
                {\includegraphics[width=\linewidth]{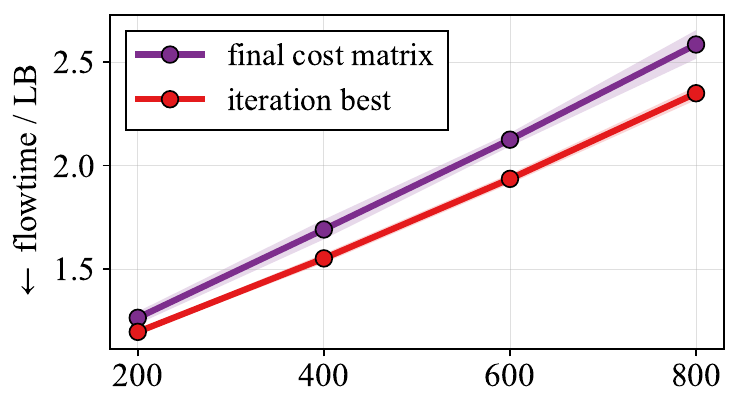}};
            \node[anchor=south, font=\small, yshift=1pt]
                at ([xshift=0.10\linewidth]img.north) {ost003d};
        \end{tikzpicture}
    \end{minipage}\hfill
    \begin{minipage}[c]{0.49\columnwidth}
        \centering
        \begin{tikzpicture}
            \node[inner sep=0] (img)
                {\includegraphics[width=\linewidth]{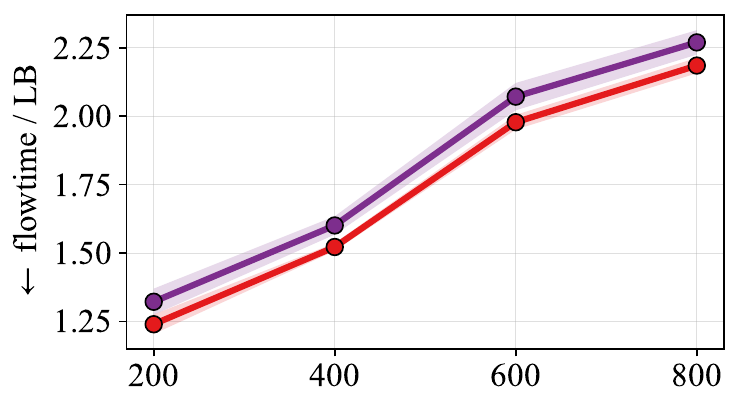}};
            \node[anchor=south, font=\small, yshift=1pt]
                at ([xshift=0.10\linewidth]img.north) {lak303d};
        \end{tikzpicture}
    \end{minipage}
    \caption{
    One-shot target assignment on the final calibrated cost matrix versus the best solution found during the iterations, in \scenHotspot.
    }
    \label{fig:final_cost_matrix}
\end{figure}
\Cref{fig:final_cost_matrix} compares the best solution found during the iterations with a one-shot assignment on the final calibrated matrix, under the same budget.
The margin is not large, but the one-shot assignment consistently falls short.
The cost matrix estimates the \emph{average} cost of each agent--target pair, whereas the best solution exploits a particular combination of pairs whose joint cost happens to be low, and such combinations are found only by evaluating candidates.
This is why \planner returns the best solution, not the assignment implied by the final cost matrix.

\section{Travel Time can Degrade Solutions}
\label{sec:loss_vs_time}
\begin{figure}[t!]
    \centering
    \begin{minipage}[c]{0.035\columnwidth}
        \centering
        \rotatebox[origin=c]{90}{\footnotesize $\leftarrow$ flowtime / LB}
    \end{minipage}\hfill
    \begin{minipage}[c]{0.31\columnwidth}
        \centering
        {\footnotesize \strut\\ \mapname{Paris\_1\_256}}\\[1pt]
        \includegraphics[width=\linewidth]{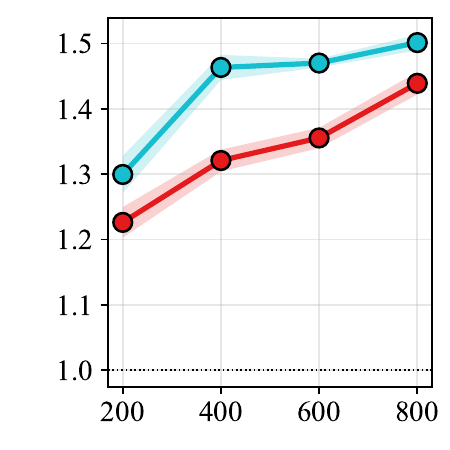}
    \end{minipage}\hfill
    \begin{minipage}[c]{0.31\columnwidth}
        \centering
        {\footnotesize \mapname{warehouse-}\\ \mapname{10-20-10-2-1}}\\[1pt]
        \includegraphics[width=\linewidth]{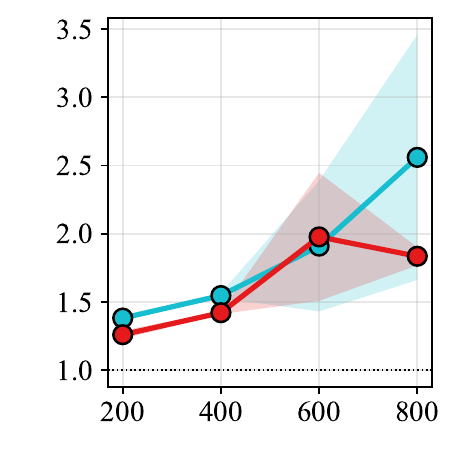}
    \end{minipage}\hfill
    \begin{minipage}[c]{0.31\columnwidth}
        \centering
        {\footnotesize \mapname{warehouse-}\\ \mapname{20-40-10-2-1}}\\[1pt]
        \includegraphics[width=\linewidth]{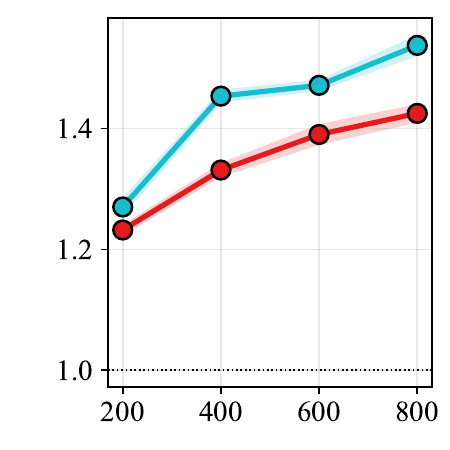}
    \end{minipage}\\[1pt]
    \hspace*{0.035\columnwidth}{\footnotesize agents}\\[3pt]
    \includegraphics[width=0.62\columnwidth]{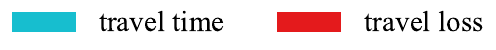}
    \caption{Comparison of calibrating with the travel time and the travel loss on \scenHotspot.}
    \label{fig:loss_vs_time}
\end{figure}
\planner calibrates its cost matrix with the travel loss rather than the travel time.
\Cref{fig:loss_vs_time} compares the two choices, demonstrating that calibrating with the travel loss reaches a better solution.
The two costs differ in how they treat the steps around the target.
Once an agent rests at its target, its travel loss stops growing, whereas its travel time keeps growing until it stops moving for good.
The observed travel time therefore also counts the steps an agent spends making way for the agents that are still moving, and these steps have little to do with the quality of the assigned pair.

\section{Experimental Details}
  \begin{table}[t!]
      \centering
      \begin{tabular}{lr}
          \toprule
          map & $d$ \\
          \midrule
          \mapname{maze-32-32-4}           & 40  \\
          \mapname{random-64-64-20}        & 43  \\
          \mapname{warehouse-10-20-10-2-1} & 76  \\
          \mapname{ost003d}                & 143 \\
          \mapname{warehouse-20-40-10-2-1} & 154 \\
          \mapname{lak303d}                & 173 \\
          \mapname{Paris\_1\_256}          & 183 \\
          \mapname{den520d}                & 187 \\
          \bottomrule
      \end{tabular}
      \caption{Distance $d$ of \scenEqdist per map.}
      \label{tab:eqdist_params}
  \end{table}
\begin{table}[t!]
    \centering
    \begin{tabular}{lrrrr}
        \toprule
        map & \multicolumn{4}{c}{hotspot radius $r$} \\
        \cmidrule(lr){2-5}
        & $n{=}50$ & $100$ & $150$ & $200$ \\
        \cmidrule(lr){2-5}
        \mapname{maze-32-32-4}           & 8  & 14 & 18 & 22 \\
        \cmidrule(lr){2-5}
        & $n{=}200$ & $400$ & $600$ & $800$ \\
        \cmidrule(lr){2-5}
        \mapname{random-64-64-20}        & 20 & 35 & 45 & 55 \\
        \mapname{warehouse-10-20-10-2-1} & 16 & 26 & 36 & 45 \\
        \mapname{ost003d}                & 20 & 30 & 45 & 60 \\
        \mapname{warehouse-20-40-10-2-1} & 21 & 32 & 41 & 48 \\
        \mapname{lak303d}                & 30 & 45 & 55 & 60 \\
        \mapname{Paris\_1\_256}          & 19 & 30 & 39 & 48 \\
        \mapname{den520d}                & 50 & 55 & 70 & 85 \\
        \bottomrule
    \end{tabular}
    \caption{Hotspot radius $r$ of \scenHotspot per map and number of agents $n$.}
    \label{tab:hotspot_params}
\end{table}
\Cref{tab:eqdist_params,tab:hotspot_params} list the scenario parameters used in our experiments.
The distance $d$ of \scenEqdist is the median pairwise shortest-path distance of the map, placing the candidate targets at a representative travel distance.
The hotspot radius $r$ of \scenHotspot grows with the number of agents so that the hotspot contains enough free vertices for the target sets.
Experiments ran on macOS 13.4, and \planner is implemented in Rust (rustc 1.96).
All randomness is controlled by explicit seeds, which are recorded in the experiment scripts of the code appendix.

\end{document}